\documentclass[10pt,twocolumn,letterpaper]{article}

\usepackage{main}
\usepackage{times}
\usepackage{epsfig}
\usepackage{graphicx}
\usepackage{amsmath}
\usepackage{amssymb}
\usepackage{multirow}
\usepackage{shortbold}
\usepackage{booktabs}
\usepackage{verbatim}
\usepackage{pifont}

\usepackage[pagebackref=true,breaklinks=true,letterpaper=true,colorlinks,bookmarks=false]{hyperref}

\iccvfinalcopy %

\def\iccvPaperID{3969} %

\newcommand{\cmark}{\ding{51}}
\newcommand{\xmark}{\ding{55}}
\newcommand{\pointx}{\mathbf{x}}

\ificcvfinal\fi

\begin{document}

\title{Collision Replay:\\ What Does Bumping Into Things Tell You About Scene Geometry?}

\author{Alexander Raistrick, Nilesh Kulkarni, David F. Fouhey\\
University of Michigan}

\maketitle
\ificcvfinal\thispagestyle{empty}\fi

\begin{abstract}
   What does bumping into things in a scene tell you about scene geometry? In this paper, we investigate the idea of learning from collisions. At the heart of our approach is the idea of 
collision replay, where we use examples of a collision to provide supervision for observations
at a past time step. We use collision replay to train a model to predict a distribution over collision time from new observation by using supervision from bumps. We learn this distribution conditioned on visual data or echolocation responses. This distribution conveys information about the
navigational affordances (e.g., corridors vs open spaces) and, as we show, can be converted
into the distance function for the scene geometry. We analyze this approach with an agent
that has noisy actuation in a photorealistic simulator.
\end{abstract}

\section{Introduction}

Suppose you bump into something. What does the collision reveal? You know your current position is on the edge of occupied space, but it is not clear what is known about other locations. If you consider where you were $k$ steps earlier, and replay this observation, there must be a $k$-step path to {\it some} obstacle, but this does not rule out a shorter path to {\it another} obstacle or noise. The goal of this paper is to use this strategy of replaying a collision ({\it Collision Replay}) to convert the local collision signal to supervision for scene geometry for other modalities like vision or sound.

\begin{figure}
    \centering
    \includegraphics[width=\linewidth]{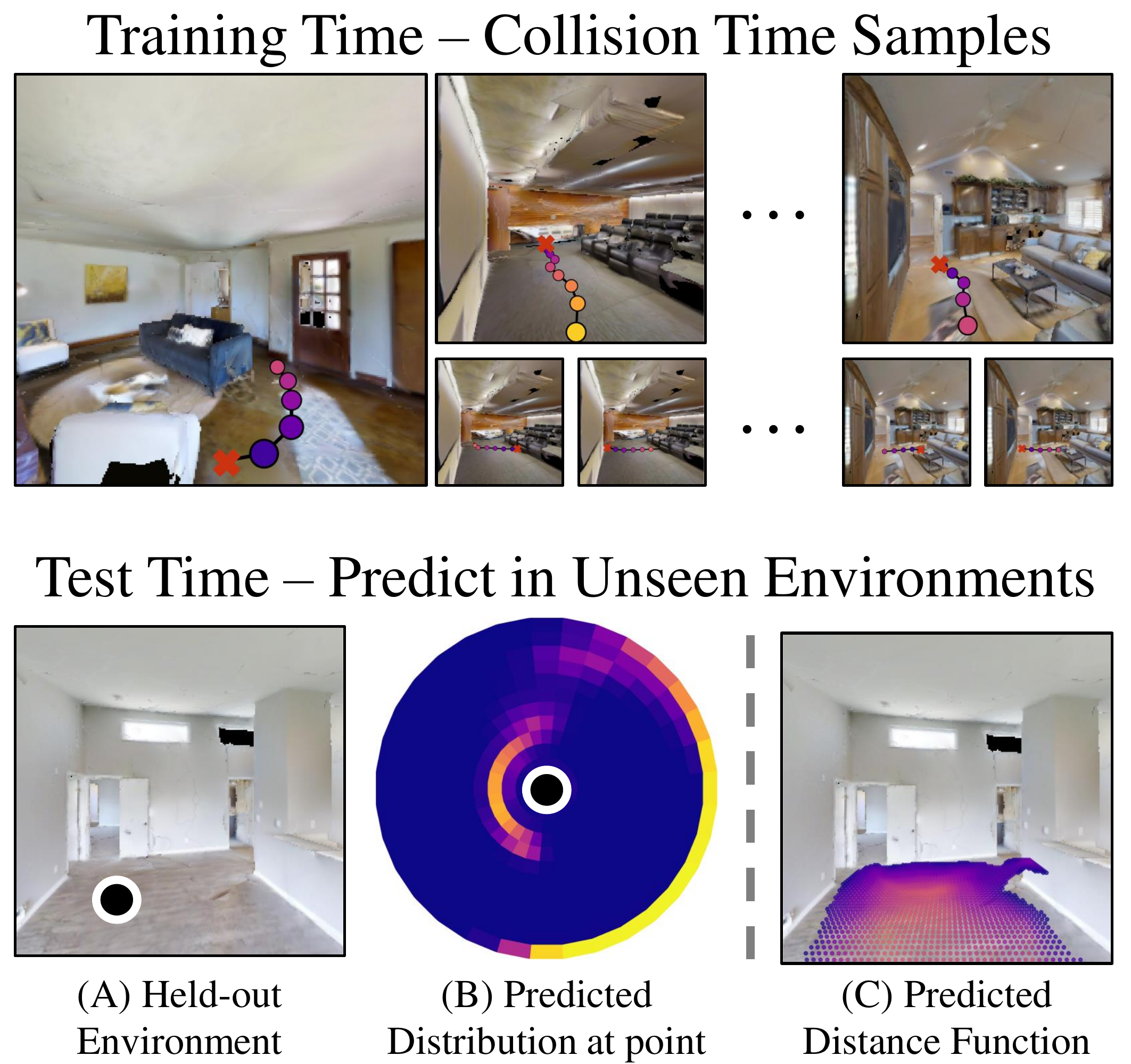}
    \caption{Our models learn from noisy samples of collisions (red crosses) in random walks. While a single collision is uninformative, multiple collisions provide signal about scene structure. At test time, given an image from an unseen environment and a point on the floor (A), our model estimates a distribution around that point conditioned on the heading angle (B) which can be converted to a  distance function (shown for all points on the floor in C).}

    \label{fig:teaser}
\end{figure}

Of course, considerable effort is usually spent in vision and robotics trying to {\it avoid} collisions. Approaches range from building detailed geometric maps~\cite{thrun2002probabilistic,hartley2000multiple,Scharstein02}, to using these maps or other signals to learn depth estimation models ~\cite{Eigen15,zhou2017unsupervised,Godard17}, to learning full-fledged navigation systems~\cite{gupta2017cognitive, yang2018visual, de2018talk} to everything in the middle. This is because a collision on its own is not particularly useful. Indeed, the most salient exception to the collision-avoidance trend~\cite{gandhi2017learning} aims to collide only as a way to learn what {\it not to do}.

While we agree that a single collision has limited use on its own, we believe that collisions have tremendous value in larger numbers and when paired with other time-locked modalities \cite{Smith2005}. The key to unlocking their value, and the core insight of our paper, is that one should model the {\it distribution} of times to collision (often studied in stochastic processes ~\cite{Feller50}), {\it conditioned} on other signals. Each individual time to collision is merely a sample from some distribution and often an over-estimate of the shortest path to an obstacle. However, there is tremendous information in the full distribution (e.g., an estimated distribution in Fig.~\ref{fig:teaser} conditioned on heading angle): the shortest plausible path is the distance to the nearest obstacle, and the layout of the scene geometry is visible in the distribution. 

We show how to operationalize this idea in Section~\ref{sec:method_model}. Given a set of random walks in a simulator, our approach learns a heading-angle conditioned function from observation features to the distribution over collision times. The approach only requires recognizing ``collisions'' (or proximity  detection) and local computations over a buffer of observations, and avoids building a full consistent map. We show generality of the idea by making predictions for multiple settings and input modalities. We predict collision time distributions from images both for remote points and at the agent's current location, and further generalize by making egocentric predictions from only sound spectrograms.

We test the value of our approach via experiments in Section~\ref{sec:experiments}. We demonstrate our approach in simulated indoor environments with realistic actuation noise, using the Habitat simulator \cite{habitat19iccv} with the Gibson~\cite{xiazamirhe2018gibsonenv} and Matterport ~\cite{Matterport3D} datasets. We primarily evaluate how well we can estimate the distance function of the scene, either projected to the floor as shown in Fig~\ref{fig:teaser} or seen from an egocentric view. Our experiments show that modeling the full distribution outperforms alternate approaches across multiple settings, and despite sparse supervision, the performance is close to that of strongly supervised methods. We additionally show that image-conditioned estimates of the distribution capture human categories of spaces (e.g., corners, hallways). 

\section{Related Work}
\label{sec:related}

The goal of this paper is to take a single {\it previously unseen} image and infer the distribution of steps to collision. We show that this distribution carries information about scene shape and local topography via its distance function. We demonstrate how we can learn this distribution from collisions of an agent executing a random walk.

Estimating scene occupancy and distance to obstacles has long been a goal of vision and robotics. Collision replay provides a principled method to derive sparse noisy training signal from random walks in any number of training environments. Models resulting from collision replay are able to predict these quantities given only a single image observation of a previously unseen environment. This separates it from the large body of work that aims to build models of a specific environment over experience, e.g., via SLAM~\cite{thrun2002probabilistic,hartley2000multiple,endres20133,cadena2016past} or bug algorithms ~\cite{mcguire2019comparative} that use cheap sensors and loop closure.  Instead, this puts our work closer to the literature in inferring the spatial layout ~\cite{lee2017roomnet, zou2018layoutnet} of a new scene. In this area, the use of collisions separates it from work that predicts floor plans from strong supervision like ground-truth floor plans~\cite{katyal2019uncertainty, shrestha2019learned, ramakrishnan2020occupancy}
or RGB-D cameras~\cite{gupta2017cognitive, chaplot2018active}. This lightweight supervision puts us most closely to work on self-supervised depth or 3D estimation, such as work using visual consistency~\cite{zhou2017unsupervised,Godard17,drcTulsiani17, watson2019self, tung2017self}. Bumps offer an alternate source of supervision that does not depend on 3D ray geometry and therefore can be used with a variety of different modalities (e.g., sound \cite{gao2020visualechoes}). We show how our method can use echo-location responses in the scene to predict a distribution over to collision. 

Our supervisory signal, time to collision, requires a few sensors and no human expertise. This is part of a trend in autonomous systems, where learning-based systems are trained on large-scale and noisy datasets of robots performing tasks often with a randomized policy. This has been successfully applied to learning to fly \cite{gandhi2017learning}, grasping~\cite{pinto2016supersizing}, poking~\cite{agrawal2016learning}, identifying when humans will collide with the robot~\cite{manglik2019forecasting}, understanding surfaces through bounce trajectories \cite{Purushwalkam19} and understanding robot terrain~\cite{kahn2020badgr}. Our work is inspired by this research and applies it by modeling the time to collision with the objects in the world (unlike \cite{manglik2019forecasting}, which modeled collision with a pedestrian). The most similar is to us is LFC~\cite{gandhi2017learning}, which predicts whether a drone will crash in the next $k$ steps from egocentric views. Our work differs in multiple ways: we use a random, not targeted, policy and output an angle- and location- conditioned distribution of steps to collision as opposed to a single  probability of collision in a fixed time horizon. Our output can be directly decoded to a distance function and is informative about scene topography. Finally, we estimate output for remote points in the scene, as opposed to just egocentric views.

Throughout, we use implicit representation learning, which has become popular for 3D tasks~\cite{xu2019disn, mescheder2019occupancy, park2019deepsdf, saito2019pifu}. We build upon the architecture in PIFu~\cite{saito2019pifu}, but rather than condition on image features to predict occupancy and color, we condition on point location, image features (or optionally features from spectrograms)  and heading to predict a distribution over collision times

\section{Method}
\label{sec:method}

The heart of our approach is modeling the distribution over time to collision conditioned on observations and heading angle. We start with an illustration of distributions of collision times in simplified scenes in (Section~\ref{sec:method_preliminary}). These collision time distributions serve as learning targets for the estimators we introduce  (Section~\ref{sec:method_model}). Having introduced the general case, we conclude by describing the particular implementations for specific settings, namely estimating distributions in scenes and egocentric views (Section~\ref{sec:method_visual}).

\begin{figure}[t]
    \centering
    \includegraphics[width=\linewidth]{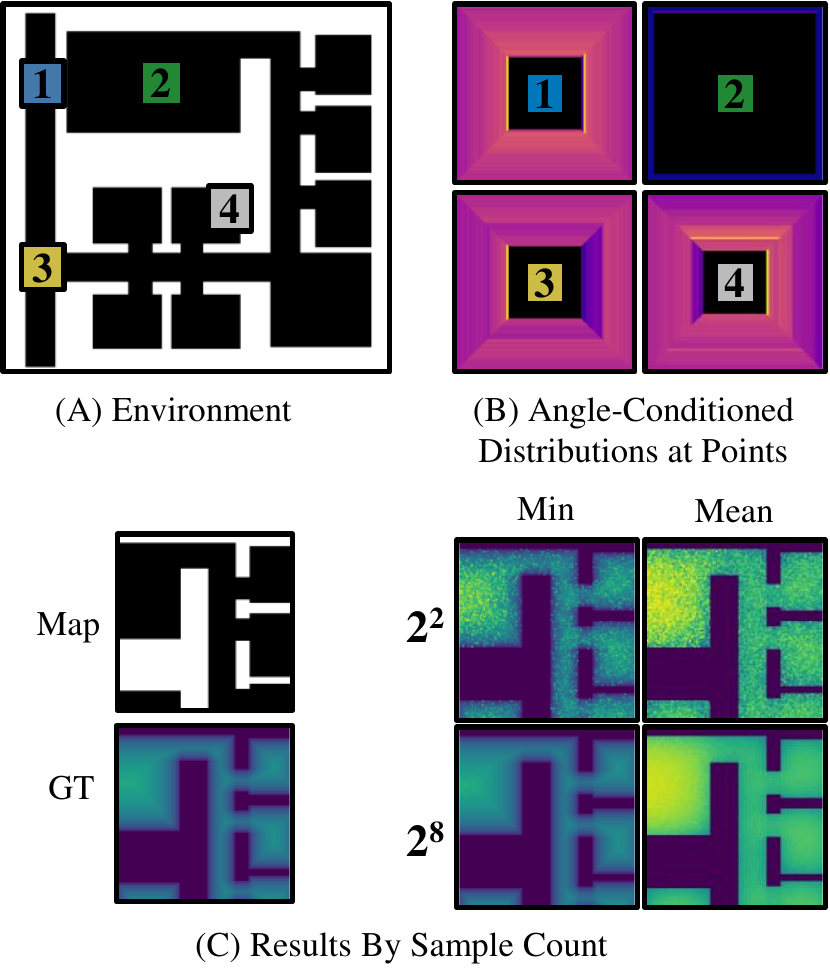}
    \caption{{\bf A simple overhead example}. A grid-cell environment with free-space shown in black {\it (A)}. We sample random walks at each grid cell for each heading $\alpha$. The distribution of hitting-times {\it (B, in log-scale)} reflects the surrounding environment. With finite samples, estimates converge fast {\it (C, showing one corner)}.}
    \label{fig:simpleworld}
     \vspace{-0.1in}
\end{figure}

\subsection{Preliminaries}
\label{sec:method_preliminary}

We begin with an illustration of collision-times in a simplified, noiseless, discrete 2D grid world shown in Fig~\ref{fig:simpleworld}. In this world, the agent is at a position $\xB$ and has heading $\alpha$ and can rotate in place left/right or take a single step forward. Let us assume the agent follows a random walk policy that uniformly selects one of forward, rotate left, or rotate right with probability ($\tfrac{1}{3}, \tfrac{1}{3}, \tfrac{1}{3}$). Then, let $T \in \{0,1,\ldots\}$ be a random variable for the number of forward steps the agent takes until it next bumps into part of the scene. The distribution over the number of forward steps until the next bump is often known as the {\it hitting time} in the study of stochastic processes and Markov chains~\cite{Feller50}. Our work aims to learn this distribution conditioned on location $\xB$ and heading $\alpha$.

If we condition on location and angle, the probability mass function over $T$, or $P_T(t|\xB,\alpha)$, tells us the likelihood for the number of steps to a collision if we start at a location and heading angle. For instance, in Fig~\ref{fig:simpleworld}, point 1, the agent is in a hallway and likely to bump into the walls if it heads East or West (seen by the sharp colors on the right and left edges of the plot). If the agent sets out at point 3, going West will surely lead to a bump while going East will take longer. 

One can convert this distribution into other quantities of interest, including the distance function to non-navigable parts of the scene as well as a floor-plan. The distance function at a location $\pointx$ is the first time $t$ which has support across any of the possible heading angles $\alpha$, or
\begin{equation}
\label{eqn:min_dist}
\textrm{DF}(\xB) = \min_{t} t \textrm{~s.t.~} \max_\alpha P_T(t|\xB,\alpha)>0,
\end{equation}
where the $\max_\alpha$ is not strictly  necessary if the agent can freely rotate and rotation does not count towards steps. This distance can be converted to a floorplan by thresholding.

Particular care is required for two modeling decisions: we assume the agent actually collides and follows a random policy at train time. In practice, one might replace collision with being close (e.g., $<10$cm) to an obstacle, measured with a proximity sensor. This enables use in environments where real collisions are not safe, and results in a model that learns a distance-to-near-collision. Additionally, useful robots do things rather than randomly walk. We use random policies to show that our system does not depend on a {\it specialized} policy, and to tie our work with theory on random walks \cite{Feller50}. Different policies change $P_T$ instead of precluding learning: the prevalence of fairly short paths provides more signal, and these may be more likely with goal-directed policies rather than ones that can spin in place. Additionally, the use of a random policy at train time does not require one use a random policy at test time.

\subsection{Modeling Collision Times in Practice}
\label{sec:method_model}

\begin{figure}[t]
    \centering
    \includegraphics[width=\linewidth]{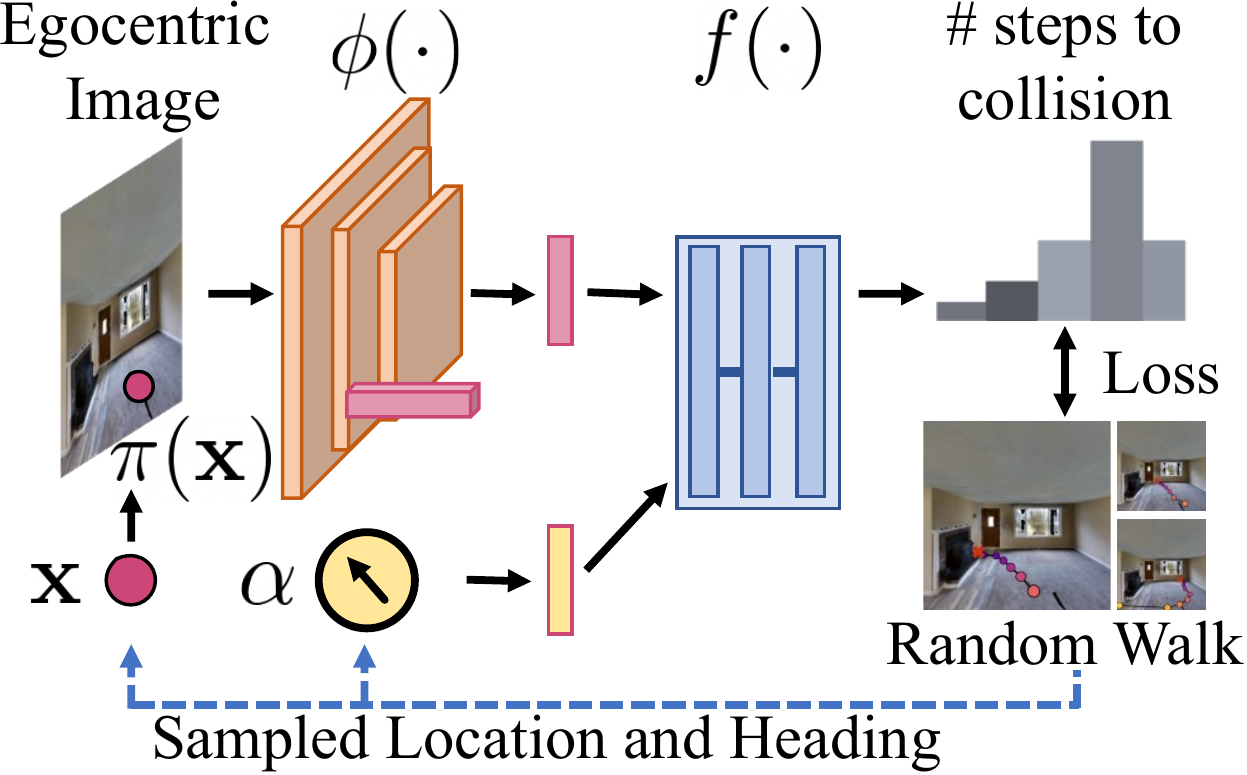}
    \caption{{\bf Overview of training pipeline}: Our approach takes a point, $\pointx$, on a random walk and predicts the distribution of times to collision conditioned on features $\phi(\pointx)$ and the heading angle $\alpha$. The prediction is supervised by the labels generated for the sampled point using collision replay.}
    \label{fig:overview}
    \vspace{-0.1in}
\end{figure}

In practice, we would like to be able to infer distributions over hitting times in new scenes, and would like to learn this from noisy samples from training scenes. We therefore learn a function $f(\phi(\pointx),\alpha)$ that characterizes
$P_T(t|\pointx,\alpha)$, for instance by giving the distribution. The first input to $f$ is image features of the location $\pointx$ (such as the output of a deep network) concatenated with the $d_{\pi}(\pointx)$, where $d_{\pi}$ is the projected depth with respect to the camera $\pi$. We denote this concatenated feature as $\phi(\pointx)$. The second input feature is the camera-relative heading angle $\alpha$. At training time, we assume each step gives a sample from the true hitting time distribution at the given location.

 We frame the problem as predicting a multinomial distribution over $k+1$ categories: $0, \ldots, k-1$ and finally $k$ or more steps. In other words $f(\phi(\pointx),\alpha)$ is a discrete distribution with with ${k+1}$ entries. 
If $f$ is learnable, one can train it to match the predictions of $f$ to $P_T(t|\pointx,\alpha)$ by minimizing the expected negative-log-likelihood over the distribution, or
$\mathbb{E}_{s \sim P_T(t|\pointx,\alpha)}\left[ -\log f(\phi(\xB),\alpha)_s \right]$. Typically, networks estimate a distribution with something like a softmax function that makes all values non-zero. Thus in practice, we modify Eqn.~\ref{eqn:min_dist} to be first time $t$ where the cumulative probability exceeds a hyperparameter $\epsilon$, or
$\min_t~\textrm{s.t.} \max_a ( \sum_{i=0}^t f(\phi(\pointx),\alpha)_i ) \ge \epsilon$. An alternate approach is predicting a scalar. For example, one could learn $g(\phi(\pointx),\alpha)$ with a regression loss like the MSE, or minimize  $\mathbb{E}_{s \sim P_T(t|\pointx,\alpha)} [(s-g(\phi(\pointx),\alpha))^2]$. this loss is minimized by the expected time to collision.

Conveniently, distributions like $P_T(t|\pointx,\alpha)$ are well-studied: in 1D, our setting corresponds to the classic Gambler's Ruin~\cite{Feller50} problem for which many results are known. A full description appears in the supplement -- the formulae are unwieldy and unintuitive. We summarize two salient results here, generating concrete numbers under the following scenario: an agent starts at cell $z=20$ in a $51$ cell world enclosed by two walls, and moves left towards the wall with $p{=}80\%$ chance and right away from the wall with $1-p$. 

First, the agent has an effectively 100\% chance of reaching $0$, but the {\it expected} time, $33$, is always an overestimate of the distance. This overestimate depends on $p$:  ($p{=}90\%$ gives 25; $p{=}45\%$ gives ${\approx}199$). Thus, networks trained with the MSE should overestimate time to collision by a factor that depends on the specific policy being executed.

Second, approximately short paths are surprisingly likely. It is true that the exact shortest path's likelihood decays exponentially with distance: there is an ${\approx}1/84$ chance of taking the shortest path at $z{=}20$. However, there is a ${\approx}1/20$ chance of a path that is at-most two more steps  and a ${\approx}1/8$ chance of a path that is at-most four more steps than the shortest paths. This suggests that networks trying to match $P_T(t|\pointx,\alpha)$ will frequently encounter samples that are close to the true shortest path. Moreover, while changing the policy  changes the distribution, the estimate of the distance function is extremely off only if train time frequency is very small for fairly short paths.

\subsection{Learning to Predict Collision Times}
\label{sec:method_visual}

\begin{figure*}[!h]
\includegraphics[width=\linewidth]{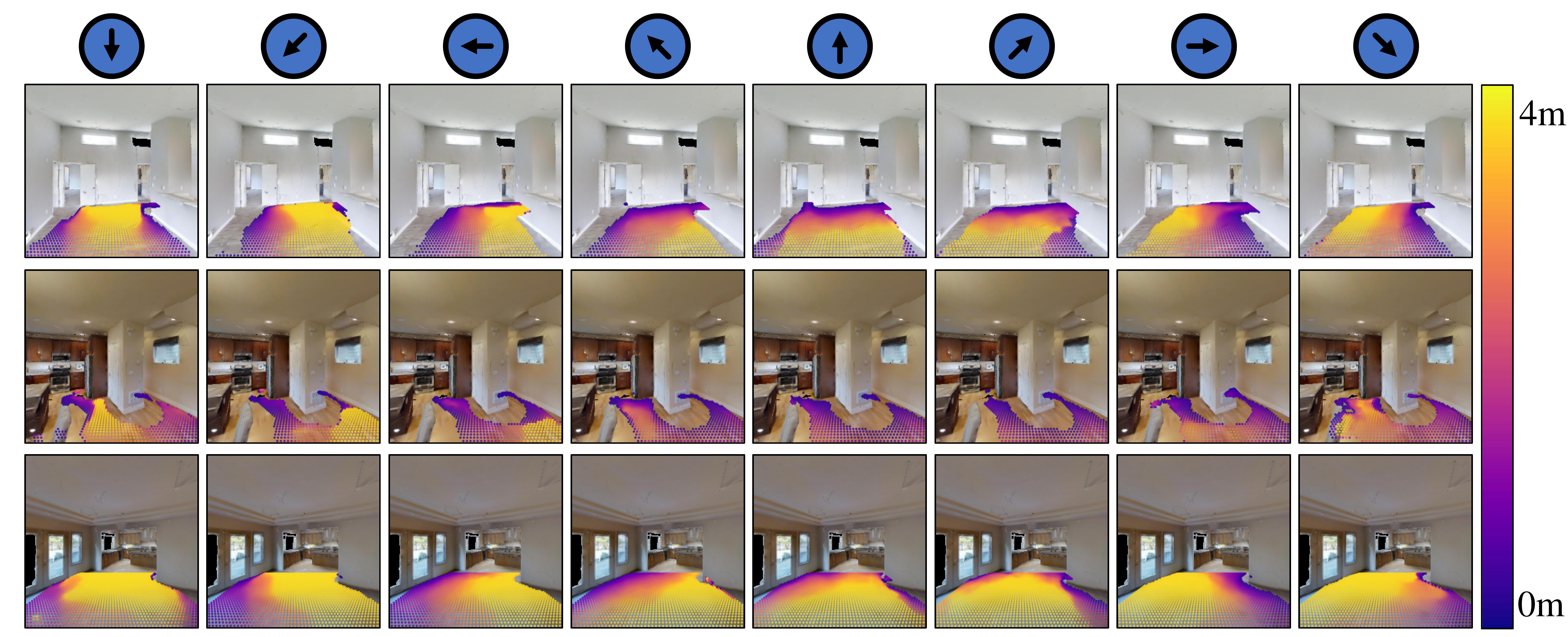}
\caption{{\bf Heading conditioned distance to collision:} Visualizations of the estimated first step with $>\epsilon$ probability of collision for grids of remote points with heading angles $\alpha$ shown as arrows. Shown in units of meters (where 1 step = 0.25m). Results are predicted by a classification model trained with noisy walks. Points are missing if the method predicted them to be occupied. } 
\vspace{-0.2in}
\label{fig:timetocollisionalpha}
\end{figure*}

At training time, we have a random walk consisting of locations $(\pointx_0, \pointx_1, \ldots, \pointx_N)$. If there is a collision at time $i$, $\pointx_i$ is labeled as zero steps to collision, $\pointx_{i-1}$ as one step, etc. This yields tuples of locations and labels that are used as training samples in a setting-dependent fashion.

\vspace{1mm}
\par \noindent {\bf Predicting Remote Locations:} Our primary setting is predicting the time to collision for a remote location on the ground-plane. This distribution can be converted into a distance function or floorplan for the scene. We predict the distribution using a PIFu \cite{pifuSHNMKL19}-like architecture shown in Fig.~\ref{fig:overview}, in which a CNN predicts image features used to condition an implicit function for each camera ray. In our case, we condition on a heading angle $\alpha$  and projected depth represented in the egocentric frame (instead of only projected depth as in \cite{pifuSHNMKL19}). Thus, $\phi(\pointx)$ is the feature map vector predicted at the location.

At train time, collision replay unwinds the agent's actuations to provide supervision at previous steps. In the case of predicting remote locations in the scene, we unwind twice to generate supervision. Suppose we consider three steps $i,j,k$ with $i<j<k$ and a collision at step $k$. We know that $j$ is $(k-j)$ steps from a collision. Then, we can project step $j$'s location into the image at step $i$ and give it the label $(k-j)$. The reasoning and projection is done assuming (incorrectly in our simulator) that the intended egomotion was executed correctly. This approach enables labeling from multiple views and through occlusion, while also not requiring explicit reconstruction. While the labels are noisy, this noise is simply propagated to the estimated distribution.

\vspace{1mm}
\par \noindent {\bf Predicting Current Locations:} To show the generality of our method, we experiment with predicting the time-to-collision distribution for the agent's {\bf current location}, conditioned on the  agent's next action and current observation. This could be a small image in a low-cost robot, sound, \cite{gao2020visualechoes} or another localized measurement like WiFi signals. We demonstrate this approach using input from two example sensors:first, a low-resolution ($32\times32$) RGB image, designed to simulate the variety of possible impoverished stimuli found on low cost robots, and second, a binaural sound spectrogram, to simulate echolocation. We input the agent's next action as it parallels the angle conditioning provided to the remote prediction model.

\vspace{1mm}
\par \noindent  {\bf Implementation Details:} 
{\it Architectures:} Both our remote location prediction and egocentric prediction network use a ResNet-18 \cite{he2016deep} backbone followed by a series of fully connected layers. In the remote prediction case, we utilise the ResNet as a Feature Pyramid Network~\cite{lin2017feature}, before decoding each feature as an implicit function in a similar style to PIFu \cite{pifuSHNMKL19}. Full architectural details appear in the supplement.
{\it Training:} We train all remote location networks for 20 epochs. The models for egocentric prediction are trained for 5 epochs due to their smaller image size. All models are optimized with Adam~\cite{Kingma2015}, following a cosine learning rate schedule with a starting value of 2e-5, a maximum value of 2e-4 after 30\% of training has elapsed, and a final value of 1e-5. We apply random horizontal image flips, as well as $N(0, 10\%)$ horizontal image shifts. In the remote setting, points are augmented with Gaussian ($\sigma = 3\mathrm{cm}$) noise in the floor plane.
{\it Classification setting:} 
Throughout, we used a maximum label of $k=10$ steps for our classification model. Labels above this value are clamped, so that a prediction of class 10
means the model predicted 10 {\it or more} steps to collision. We found this $k$ sufficient to capture the interesting parts of most scenes. 
\section{Experiments}
\label{sec:experiments}

We now describe a series of experiments that aim to investigate the effectiveness of the proposed approach. These experiments are conducted in a standard simulation environment, including results with actuation noise. After explaining the setup {\bf (Section~\ref{sec:exp_setup}),} we introduce a series of experiments. We first evaluate predicting on remote points 
{\bf (Section ~\ref{sec:exp_remote})}, comparing our approach to a variety of alternative architectures and upper bounds. We then analyze what the predicted time-to-collision distributions represent {\bf (Section~\ref{sec:exp_analysis})}. Finally, we demonstrate the generality of the idea by applying it to egocentric views{\bf (Section~\ref{sec:exp_egocentric})}.

\subsection{Environment and Dataset}
\label{sec:exp_setup}

All experiments use random walk data collected from the Habitat simulator ~\cite{habitat19iccv}. We use the Gibson Database of 3D Spaces ~\cite{xiazamirhe2018gibsonenv} for all experiments except one, the egocentric sound comparison in Table \ref{tab:egocentric_sound}, which uses Matterport3D \cite{Matterport3D} in order to support sound simulation. We hold out entire environments according to the splits provided by each dataset so that our training, validation and testing sets are entirely independent. In each environment, we perform 10 random walks of 500 steps each, starting at a random location for each walk. At each time-step, we collect the agent's collision state,  sensor data (egocentric images, or optionally sound spectrograms), and ego-motion.

\vspace{1mm}
\par \noindent {\bf Agent and Action Space:} We model the agent as a cylinder with a radius of 18cm, a forwards facing camera 1.5 meters above the ground plane, and a binary collision sensor. At each timestep, the agent selects one of four actions. We follow Active Neural Slam \cite{chaplot2020learning}: {\it move forward} (25cm), {\it turn left} ($-10^\circ$), {\it turn right} ($10^\circ$). We also give the agent a {\it turn around} ($180^\circ$) action for use after a collision. When noise is active, we use the noise model from \cite{chaplot2020learning}, which is a set of Gaussian Mixture Models extracted from a LoCoBot \cite{murali2019pyrobot} agent; we assume the {\it turn around} action is distributed like {\it turn left} but with a mean rotation of 180. 
When collecting data for egocentric prediction experiments, we increase the mean rotation of the turn left and turn right actions to $-45^\circ$ and $+45^\circ$ respectively. This means the network is not forced to make fine-grained distinctions between angles given a small image while also ensuring the space of possible turn angles covers the $90^\circ$ field of view of the camera.

\vspace{1mm}
\par \noindent {\bf Policy:} Our agent conducts a random walk by selecting an action from the set ({\it move forward}, {\it turn left}, {\it turn right}) with probabilities (60\%, 20\%, 20\%) so long as it did not collide with an obstacle in the previous time step. Upon collision, the agent executes a {\it turn around} action to prevent getting stuck against obstacles and walls. As explained in \ref{sec:method_preliminary}, this random walk policy is chosen for its simplicity, and the method does not depend on this policy. 

\subsection{Predicting Remote Time-To-Collision} 
\label{sec:exp_remote}

\begin{figure}[t]
\includegraphics[width=\linewidth]{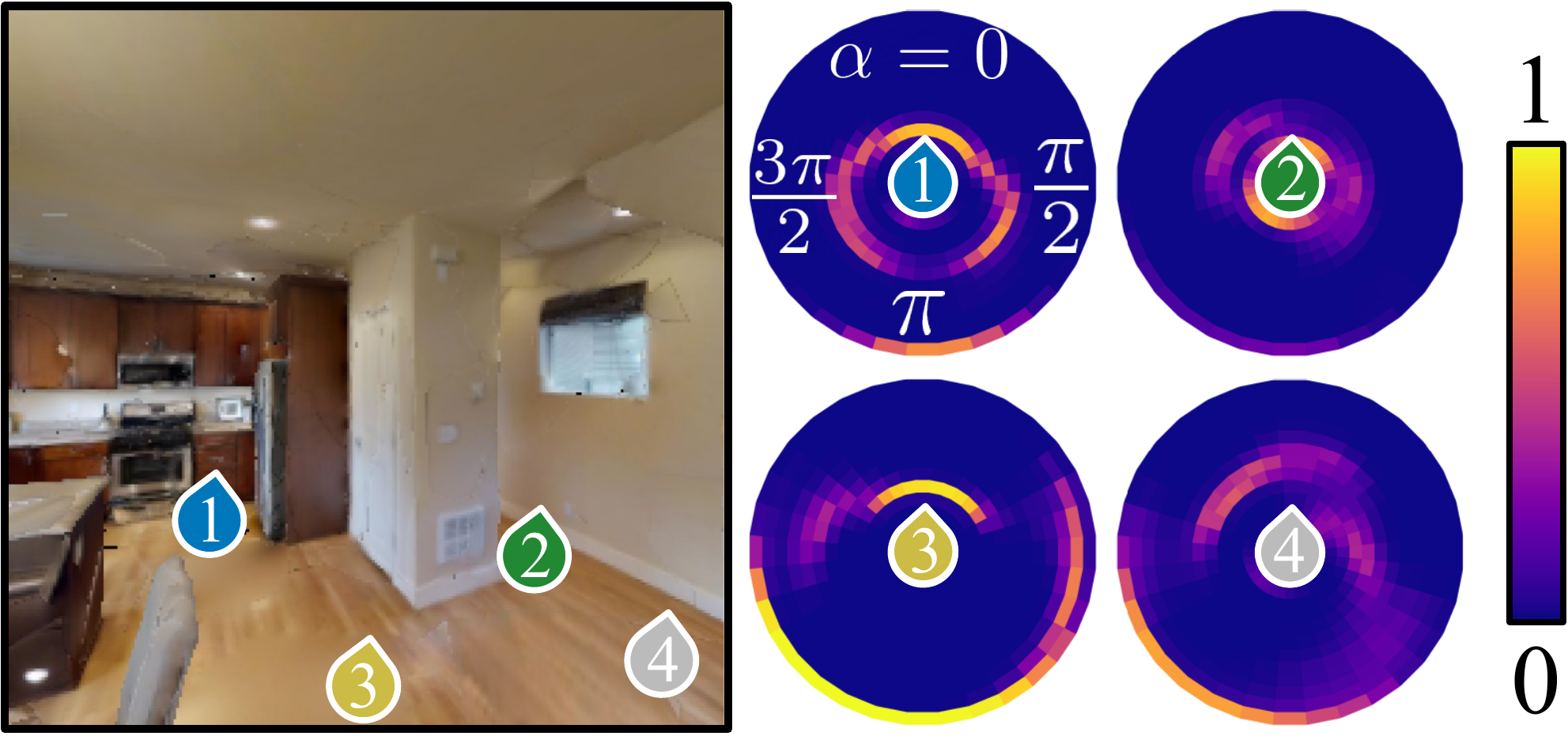}
\caption{{\bf Remote prediction for selected points:} Examples of $P(t|\alpha)$ for selected test set points. We plot the heading angle as the angle axis of a polar plot, while the radius is proportional to steps to collision. At point 3, a collision is likely soon going straight; at point 2, it is more likely going left or right.}
\vspace{-0.2in}
\label{fig:distribution_remote}
\end{figure}

\begin{figure*}
\includegraphics[width=\linewidth]{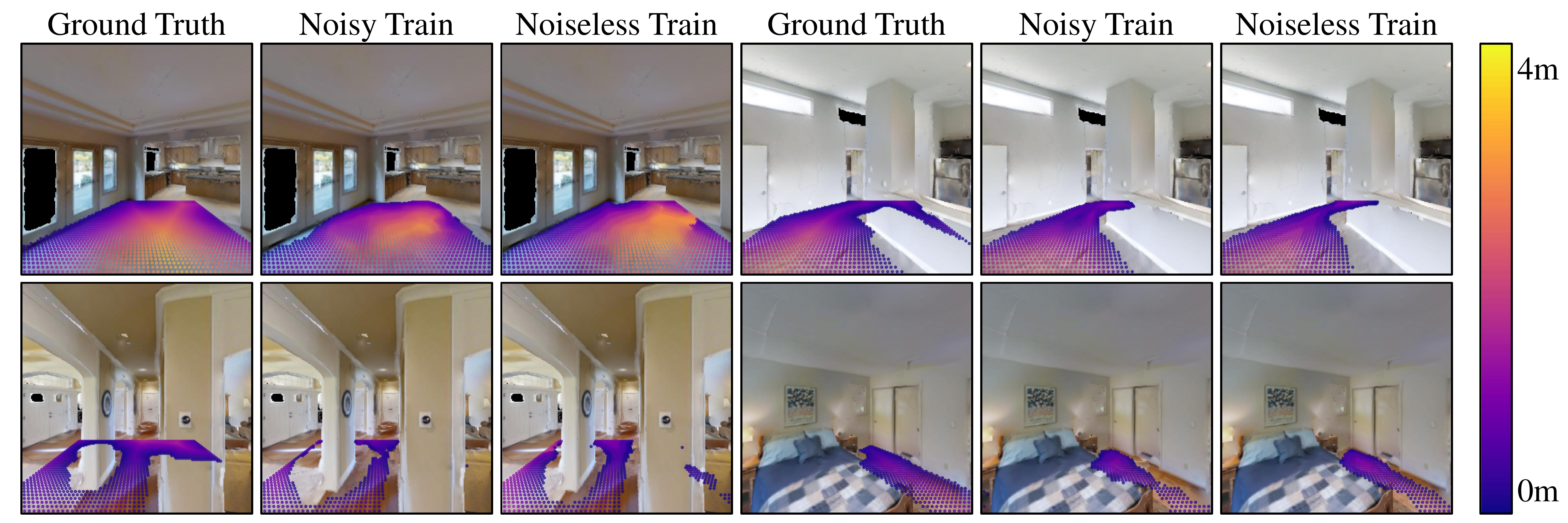}
\caption{{\bf Extracted 2D distance functions:} Examples of 2D scene distance functions extracted from (left) the simulator navigation-mesh; (middle) a model trained on noisy random walks; and (right) a model trained on noise-free random walks. The center the rooms are correctly predicted to have higher distances (marked by lighter coloring). As in Fig.~\ref{fig:timetocollisionalpha}, points are absent if they are predicted occupied.}
\label{fig:distance_function}
\vspace{-0.1in}
\end{figure*}

\vspace{1mm}
\par \noindent {\bf Qualitative Results:} In Fig.~\ref{fig:timetocollisionalpha} we show the maps of collision times for specific heading angles for all points within a 4 m x 4 m grid on the floor in front of the camera. We observe that the model correctly varies the time-to-collision such that lower values are predicted at points where the current heading value $\alpha$ faces a nearby obstacle, and higher values are predicted where the heading value faces into open areas. The model is able to infer some regions where the distance function is nonzero despite that the floor plane itself is occluded by obstacles. In Fig.~\ref{fig:distance_function} we visualize the distance function obtained from our angle conditioned outputs with both noisy and (for comparison) noise-free egomotion training. The model produces a distance function considering the closest object in any direction to each point in the scene, rather than the distance in one direction of interest as in  Fig.~\ref{fig:timetocollisionalpha}. The distance function attains high values in the middle of open spaces, and smaller values in enclosed hallways and near obstacles.

Our model is capable on generating the distribution of steps to collision conditioned on the input heading angle $\alpha$. In Fig.~\ref{fig:distribution_remote} we visualize this distribution for four points of interest by varying the heading angle. We show  polar plots where the radius of the ring denotes the distance from the point, and the color denotes the probability. At point $1$ collisions are likely for all directions except reverse, and imminent directly ahead. Similarly at point $3$, the upcoming pillar appears as shown as high probability ahead.

\vspace{1mm} 
\par \noindent {\bf Comparisons:} We compare against various approaches and ablations. For fairness, unless otherwise specified, we use an identical ResNet18 ~\cite{he2016deep} and PIFu ~\cite{pifuSHNMKL19} backbone as described in Section \ref{sec:method_visual}, changing only the size of the outputs to accommodate different targets. All outputs can be converted into a predicted floorplan and a scene distance function.
Floorplans are with a distance transform. 
\par \noindent
({\it Regression-L1/Regression-L2}): We use smoothed L1 loss (or L2 loss) to regress a scalar steps to collision value for each point.
\par \noindent 
({\it Free Space Classification}): We discretize outputs into colliding and non-colliding points. We train a binary classification model which directly predicts this outcome at each trajectory time-step, thereby learning a free-space predictor.

\par \noindent
({\it Supervised Encoder-Decoder}): We predict floor plans using strong supervision in the form of dense ground truth floor-plans collected from the environments. This baseline serves as an upper-bound for our method because our method uses sparser and weaker supervision.

\par \noindent 
({\it Analytical Depthmap Projection}): We use simulator depth to predict free-space for visible regions in the scene by analytically projecting the depth maps on the floor plane. 

\begin{table}[t]
    \centering
    \caption{Quantitative results for distance functions and floorplans.}
    \label{tab:remote}
    \begin{tabular}{@{~}l@{~~}c@{~~}c@{~~}c@{~~}c@{~~}c@{~~}c@{~}} \toprule
        &  & & \multicolumn{3}{c}{Distance Function} & Floor
        \\
       &  Angle & Noise & MAE & RMSE & $\% \le \delta$ & IoU 
        \\
        \midrule
    Classification & \xmark & \cmark & 0.16  & 0.31 & 0.77 & \bf 0.49
    \\
    Regression-L1 & \xmark & \cmark & 0.25 & 0.38 & 0.66 & 0.47
    \\
    Regression-L2 & \xmark & \cmark & 0.25 & 0.38 & 0.66 & 0.47 
    \\
    \midrule
    Classification & \cmark & \cmark & \bf 0.11 & \bf 0.21 & \bf 0.83 & 0.47
    \\
    Regression-L1 & \cmark & \cmark & 0.13 & 0.24 & 0.79 & 0.45
    \\
    Regression-L2 & \cmark & \cmark & 0.14 & 0.24 & 0.79 & 0.47
    \\
    Free-Space & \xmark & \cmark & 0.13 & 0.23 & 0.82 & 0.52
    \\
    \midrule
    Classification & \cmark & \xmark & \bf0.10 & \bf 0.20 & \bf 0.86 & 0.54
    \\
    Regression-L1 & \cmark & \xmark & 0.11 & 0.22 & 0.83 & 0.49
    \\
    Regression-L2 & \cmark & \xmark & 0.12 & 0.22 & 0.83 & 0.53
    \\
    \midrule
    Supervised & - & \xmark & \bf 0.08 & \bf 0.19 & \bf 0.90 & \bf 0.66
    \\
    Depthmap & - & \xmark & 0.09 & 0.20 & 0.87 & 0.57
    \\
    \bottomrule
    \end{tabular}
    \vspace{-0.1in}
\end{table}

\begin{figure*}[t]
\includegraphics[width=\linewidth]{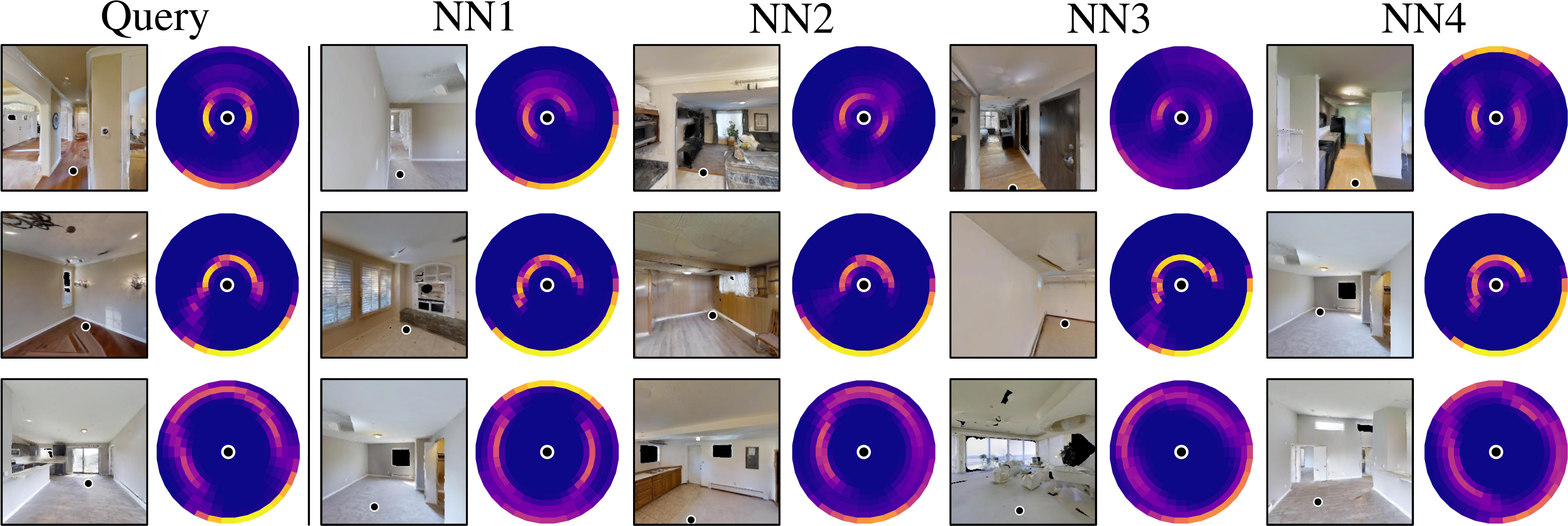}
\caption{{\bf Nearest neighbor lookup results:} Examples of nearest neighbors using Eqn.~\ref{eqn:distance} on predicted hitting time distributions, and selecting only one per-scene. Top: passing through a doorway; Middle: corner of a room; Bottom: center of room. }
\vspace{-0.1in}
\label{fig:distancefunctions}
\end{figure*}

\vspace{1mm}
\par \noindent {\bf Metrics:} We evaluate each timestep on a $64 \times 64$ grid of points covering a $4$m $\times$ $4$m square in front of the agent. We compute both the distance to the navigation mesh and the whether the point is free space. We use these labels to compute multiple metrics. 
To evaluate distance functions, we compare the ground-truth $y_i$ and prediction $\hat{y}_i$ and report: 
mean absolute error (MAE), $\frac{1}{N} \sum_i |y_i - \hat{y}_i|$; 
root-mean squared error (RMSE), $\sqrt{\frac{1}{N} \sum_i (y_i - \hat{y}_i)^2}$; and
percent within a threshold ($\delta$) $\frac{1}{N} \sum_i 1(|y_i - \hat{y}_i| < \delta)$. Throughout, we use $\delta = 0.25$m, the average step distance.
We evaluate floorplans with {\it Floorplan IoU}, the intersection-over-union between predicted and ground truth floorplans.

\vspace{1mm}
\par \noindent {\bf Quantitative Results:} We report results in Table~\ref{tab:remote}, with distance measured in meters. In each both noise-free and noisy settings, the approach of modeling with classification produces better distance functions compared to regression approaches. L1 and L2 regression, which are equivalent to estimating the median or average time to collision respectively, under-perform classification, and we find analytically that they are usually an over-prediction. Floorplan IoU does not differentiate between distances higher than the threshold, and so most approaches do well on it. Unsurprisingly, training to predict free-space via collision replay outperforms systems that obtain the floorplan estimate by indirect analysis of the distance function. Nonetheless, most methods trained via collision replay produce results that are close to those obtained by the strongly supervised approach (using RGB input) or the analytical depthmap projection (using RGBD input). Beyond particular metrics in estimating floorplan distance functions, we see advantages to the rich representation of collision time distributions, and analyze them next.

\subsection{Analysis of Learned Distributions}
\label{sec:exp_analysis} 
Qualitatively, we find that the angle-conditioned steps to collision distributions correlate with underlying scene geometry: the distribution in a hallway is different compared to one in the center of the room. We thus analyze whether the distributions contain this information.

Given a point in an image we aim to evaluate its similarity to other points in terms of it's steps to collision distribution. We start with distributions  $P_1(t|\pointx_1,\alpha_1)$ and $P_2(t|\pointx_2,\alpha_2)$ conditioned on two points $\pointx_1$ and $\pointx_2$ and compute dissimilarity using a function that also aligns the relative angle between them, followed by the Jensen-Shannon divergence (JSD), or:
\begin{equation}
\label{eqn:distance}
\min_{\theta} \textrm{JSD}(P_1(t|\pointx_1,\alpha),P_2(t|\pointx_2,(\alpha+\theta \textrm{~mod~} 2\pi))).
\end{equation}

\vspace{1mm}
\noindent {\bf Qualitative Results:} We find nearest neighbors for points using Eqn.~\ref{eqn:distance}. The top row of Fig~\ref{fig:distancefunctions}, shows that a query point in a hallway returns nearest neighbors that are also doorways or hallways with the similar distributions of $P(t|\pointx,\alpha)$. Similarly, the second row, a query point in a corner yields the corners of other similarly large and open rooms.

\vspace{1mm}
\noindent {\bf Quantitative Results:} 
We quantify this trend by annotating a subset of points in the test set with one of 5 classes: Open Space, Wall, Hallway, Concave Corner, or Crowded Space. Many points fall between or outside the categories, so we only evaluate on the examples in which 7 out of 9 annotators agreed. This results in a dataset of 1149 points with strong consensus. We evaluate how well Eqn.~\ref{eqn:distance} predicts that two locations share the same class, measuring performance with AUROC ($0.5$ is chance performance). Our method obtains an AUROC of 0.72 distinguishing all 5 labels. Some pairs of labels are easier to distinguish than others: Open-vs-Crowded Space has an AUROC of 0.87 and Wall-vs-Corner has an AUROC of 0.55. 

\subsection{Predicting Egocentric Time-To-Collision}
\label{sec:exp_egocentric}

Finally, we use collision replay to provide supervision for egocentric observations. We further demonstrate generality with two input modalities: small ($32 \times 32$) images, and binaural echolocation spectrograms. Given one such input, we predict collision time conditioned on what one does next ({\it turn left}, {\it move forwards}, {\it turn right}).

\begin{figure}
\includegraphics[width=\linewidth]{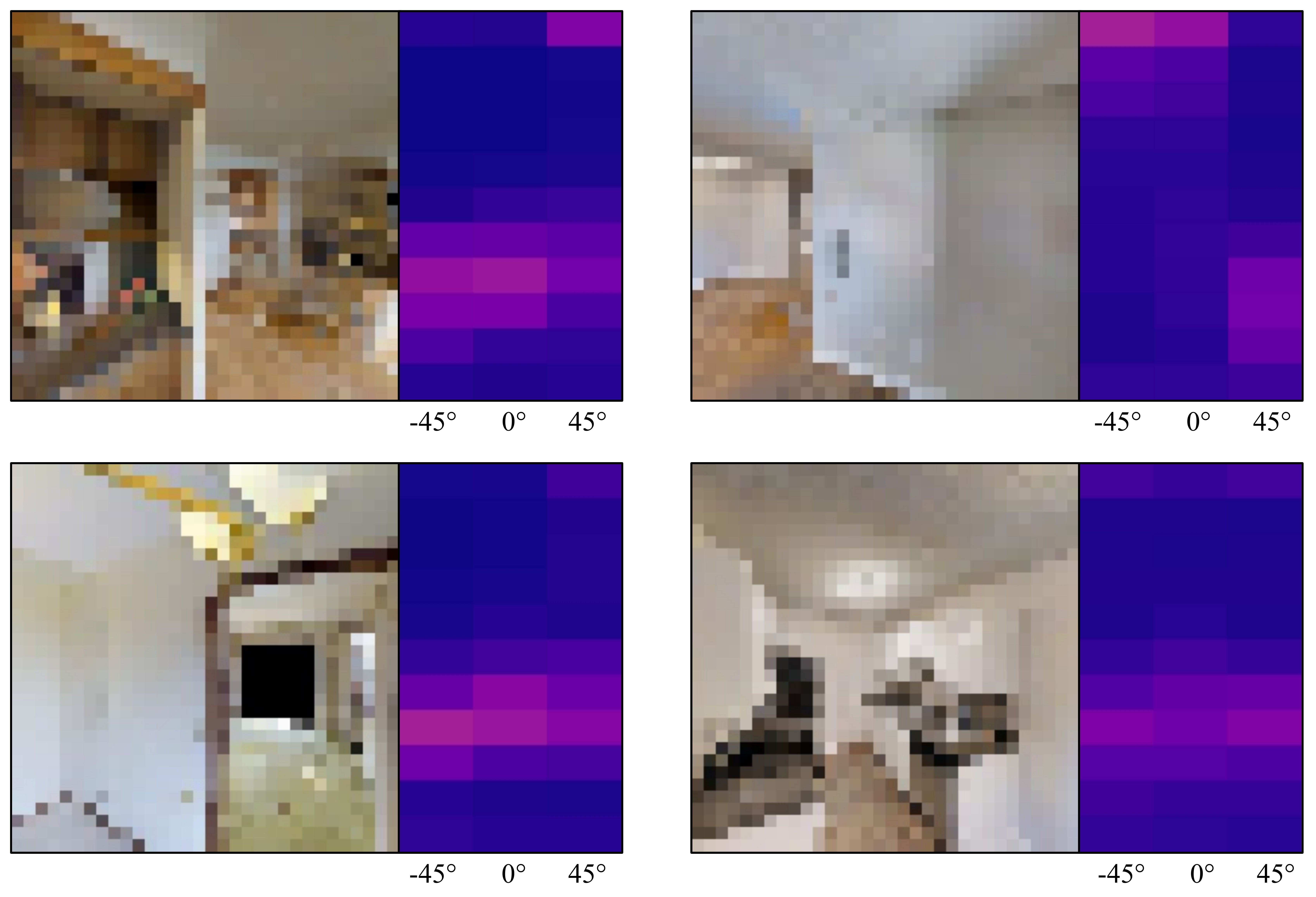}
\caption{{\bf Egocentric test predictions using 32x32 images:} The center column is the distribution with no rotation; the left/right show the distribution for turning left/right by $45^{\circ}$ }
\label{fig:egocentric}
\vspace{-0.1in}
\end{figure}

\vspace{1mm}
\par \noindent {\bf Qualitative Results:} We show qualitative results for the egocentric vision task in Fig.~\ref{fig:egocentric}. Spectrogram results are in the supplement. The predicted steps-to-collision distribution yields a rough depthmap: in Fig.~\ref{fig:egocentric}(top-left), the distribution suggests that right side of the scene is further away than ahead or to the left. Additionally, the model predicts a bi-modal distribution in Fig.~\ref{fig:egocentric}(bottom-left),  revealing uncertainty about whether the random walk policy would go through the doorway or collide with it.

\begin{table}[]
    \centering
    \caption{Quantitative results for visual egocentric prediction}
    \begin{tabular}{lccc} \toprule
        & \multicolumn{3}{c}{Distance Function} 
        \\
        & MAE & RMSE & $\% \le t$ 
        \\ \midrule
        Regression-L1 & 0.96 & 1.14 & 0.21
        \\
        Regression-L2 & 0.96 & 1.18 & 0.20
        \\
        Classification & \textbf{0.58} & \textbf{0.79} & \textbf{0.36}
        \\
        \bottomrule
    \end{tabular}
    \vspace{-0.1in}
    \label{tab:ego_quantitative}
\end{table}

\vspace{1mm}
\par \noindent {\bf Quantitative Results (Visual Input):}
We compare with regression approaches from the previous section in Table~\ref{tab:ego_quantitative}. Modeling the distribution with classification outperforms regression, and regression systematically overestimates ($90\%$ of cases compared to $60\%$ for classification).

The learning task used in Learning to Fly by Crashing (LFC)~\cite{gandhi2017learning} is a special case of our method since it models the probability that $t<k$ for a fixed $k$ as opposed to our full distribution over $t$. We note that comparisons are difficult since LFC is aimed at producing a policy as opposed to scene structure, only applies in the egocentric setting, and also uses a specialized policy to collect data. If we compare our approach on the LFC task, predicting the binary outcome ``would the agent crash within k steps'', our approach outperforms LFC on F1 score for all discretizations ($k=8$: $0.66$-vs-$0.62$ and $k=2$: $0.86$-vs-$0.85$), likely due to multi-task learning effects. When comparing on our task, there is not a principled way to convert LFC's binary probability to a distance function, and results are highly dependent on the binarization threshold $k$: LFC can produce reasonable results with large $k$  (RMSE for $k=8$ is $0.84$) but small ones fare worse (RMSE for $k=2$ is $1.46$). Full experiments are in the supplement. 

\vspace{1mm}

\par \noindent {\bf Quantitative Results (Echolocation):} In the echolocation setting we compare a model with sound input to the best egocentric vision model. We train both on Matterport3D \cite{Matterport3D}, which is necessary in order to collect echo spectrograms from Soundspaces ~\cite{chen20soundspaces}, which does not currently support Gibson. The vision model is identical to the classification model in Table \ref{tab:ego_quantitative}; the sound model only replaces the ResNet-18 backbone with an Audio-CNN provided by the Soundspaces baseline code. We compare the performance of these models in Table \ref{tab:egocentric_sound}.

Sound simulation relies on pre-computed Room Impulse Response (RIR) files, which are only provided for a grid of locations and orientations. We snap to the nearest orientation, then apply Gaussian RBF interpolation on computed spectrogram outputs based on the agent location. This yields an approximate echo response for a continuous agent pose, enabling the use of the same continuous agent policy, and outperformed other methods of interpolation we tried.

\begin{table}[]
    \centering
    \caption{Quantitative results on~\cite{Matterport3D} with different inputs (vision and sound) showing the generality of collision replay.}
    \label{tab:egocentric_sound}
    \begin{tabular}{lccc} \toprule
        Input  & \multicolumn{3}{c}{Distance Function} 
        \\
         & MAE & RMSE & $\% \le t$ 
        \\ \midrule
        $32 \times 32$ Image & 0.64 & 0.86 & 0.33
        \\
        Binaural Spectrogram & \textbf{0.58} & \textbf{0.79} & \textbf{0.36}
        \\
        \bottomrule
    \end{tabular}
    \vspace{-0.1in}
\end{table}

Table \ref{tab:egocentric_sound} shows that echolocation outperforms the 32x32 image input. We attribute this to the rich information in the audio spectrograms - the return time of emitted sound is closely related to the underlying depth map of the scene \cite{gao2020visualechoes}, so it provides more signal than coarse color data. Additionally, while the sound responses are directional, sound bounces encode information about space all around the agent, which is useful since the distance function may depend on objects not visible in the camera's field of view.

\section{Conclusions and Discussion}
This paper has introduced Collision Replay, which enables using bumps to learn to estimate scene geometry in new scenes. We showcase two applications of our method using sound and vision as input, but see potential in a variety of other settings, ranging from manipulation in occluded regions to converting travel time to scene maps.

\vspace{1mm}
\noindent {\bf Acknowledgments:} This work was supported by the DARPA Machine Common Sense Program. NK was supported by TRI. Toyota Research Institute (``TRI'') provided
funds to assist the authors with their research but this article solely reflects the
opinions and conclusions of its authors and not TRI or any other Toyota entity

{\small
\bibliographystyle{ieee_fullname}
\bibliography{egbib}

\begin{thebibliography}{10}\itemsep=-1pt

\bibitem{agrawal2016learning}
Pulkit Agrawal, Ashvin Nair, Pieter Abbeel, Jitendra Malik, and Sergey Levine.
\newblock Learning to poke by poking: Experiential learning of intuitive
  physics.
\newblock {\em arXiv preprint arXiv:1606.07419}, 2016.

\bibitem{cadena2016past}
Cesar Cadena, Luca Carlone, Henry Carrillo, Yasir Latif, Davide Scaramuzza,
  Jos{\'e} Neira, Ian Reid, and John~J Leonard.
\newblock Past, present, and future of simultaneous localization and mapping:
  Toward the robust-perception age.
\newblock {\em IEEE Transactions on robotics}, 32(6):1309--1332, 2016.

\bibitem{Matterport3D}
Angel Chang, Angela Dai, Thomas Funkhouser, Maciej Halber, Matthias Niessner,
  Manolis Savva, Shuran Song, Andy Zeng, and Yinda Zhang.
\newblock Matterport3d: Learning from rgb-d data in indoor environments.
\newblock {\em International Conference on 3D Vision (3DV)}, 2017.

\bibitem{chaplot2020learning}
Devendra~Singh Chaplot, Dhiraj Gandhi, Saurabh Gupta, Abhinav Gupta, and Ruslan
  Salakhutdinov.
\newblock Learning to explore using active neural slam.
\newblock In {\em International Conference on Learning Representations (ICLR)},
  2020.

\bibitem{chaplot2018active}
Devendra~Singh Chaplot, Emilio Parisotto, and Ruslan Salakhutdinov.
\newblock Active neural localization.
\newblock {\em arXiv preprint arXiv:1801.08214}, 2018.

\bibitem{chen20soundspaces}
Changan Chen, Unnat Jain, Carl Schissler, Sebastia Vicenc~Amengual Gari, Ziad
  Al-Halah, Vamsi~Krishna Ithapu, Philip Robinson, and Kristen Grauman.
\newblock Soundspaces: Audio-visual navigaton in 3d environments.
\newblock In {\em ECCV}, 2020.

\bibitem{de2018talk}
Harm de Vries, Kurt Shuster, Dhruv Batra, Devi Parikh, Jason Weston, and Douwe
  Kiela.
\newblock Talk the walk: Navigating new york city through grounded dialogue.
\newblock {\em arXiv preprint arXiv:1807.03367}, 2018.

\bibitem{Eigen15}
David Eigen and Rob Fergus.
\newblock Predicting depth, surface normals and semantic labels with a common
  multi-scale convolutional architecture.
\newblock In {\em ICCV}, 2015.

\bibitem{endres20133}
Felix Endres, J{\"u}rgen Hess, J{\"u}rgen Sturm, Daniel Cremers, and Wolfram
  Burgard.
\newblock 3-d mapping with an rgb-d camera.
\newblock {\em IEEE transactions on robotics}, 30(1):177--187, 2013.

\bibitem{Feller50}
William Feller.
\newblock {\em An Introduction to Probability Theory and Its Applications}.
\newblock Wiley and Sons, New York, 1950.

\bibitem{gandhi2017learning}
Dhiraj Gandhi, Lerrel Pinto, and Abhinav Gupta.
\newblock Learning to fly by crashing.
\newblock In {\em 2017 IEEE/RSJ International Conference on Intelligent Robots
  and Systems (IROS)}, pages 3948--3955. IEEE, 2017.

\bibitem{gao2020visualechoes}
Ruohan Gao, Changan Chen, Ziad Al-Halab, Carl Schissler, and Kristen Grauman.
\newblock Visualechoes: Spatial image representation learning through
  echolocation.
\newblock In {\em ECCV}, 2020.

\bibitem{Godard17}
Cl{\'{e}}ment Godard, Oisin {Mac Aodha}, and Gabriel~J. Brostow.
\newblock Unsupervised monocular depth estimation with left-right consistency.
\newblock In {\em CVPR}, 2017.

\bibitem{gupta2017cognitive}
Saurabh Gupta, James Davidson, Sergey Levine, Rahul Sukthankar, and Jitendra
  Malik.
\newblock Cognitive mapping and planning for visual navigation.
\newblock In {\em Proceedings of the IEEE Conference on Computer Vision and
  Pattern Recognition}, pages 2616--2625, 2017.

\bibitem{hartley2000multiple}
R Hartley and A Zisserman.
\newblock Multiple view geometry in computer vision, cambridge uni.
\newblock {\em Pr., Cambridge, UK}, 2000.

\bibitem{He2015}
Kaiming He, Xiangyu Zhang, Shaoqing Ren, and Jian Sun.
\newblock Deep residual learning for image recognition.
\newblock {\em arXiv preprint arXiv:1512.03385}, 2015.

\bibitem{he2016deep}
Kaiming He, Xiangyu Zhang, Shaoqing Ren, and Jian Sun.
\newblock Deep residual learning for image recognition.
\newblock In {\em CVPR}, 2016.

\bibitem{kahn2020badgr}
Gregory Kahn, Pieter Abbeel, and Sergey Levine.
\newblock Badgr: An autonomous self-supervised learning-based navigation
  system.
\newblock {\em arXiv preprint arXiv:2002.05700}, 2020.

\bibitem{katyal2019uncertainty}
Kapil Katyal, Katie Popek, Chris Paxton, Phil Burlina, and Gregory~D Hager.
\newblock Uncertainty-aware occupancy map prediction using generative networks
  for robot navigation.
\newblock In {\em 2019 International Conference on Robotics and Automation
  (ICRA)}, pages 5453--5459. IEEE, 2019.

\bibitem{Kingma2015}
Diederik~P. Kingma and Jimmy Ba.
\newblock Adam: {A} method for stochastic optimization.
\newblock In Yoshua Bengio and Yann LeCun, editors, {\em ICLR}, 2015.

\bibitem{lee2017roomnet}
Chen-Yu Lee, Vijay Badrinarayanan, Tomasz Malisiewicz, and Andrew Rabinovich.
\newblock Roomnet: End-to-end room layout estimation.
\newblock In {\em Proceedings of the IEEE International Conference on Computer
  Vision}, pages 4865--4874, 2017.

\bibitem{fpn}
T. {Lin}, P. {Dollár}, R. {Girshick}, K. {He}, B. {Hariharan}, and S.
  {Belongie}.
\newblock Feature pyramid networks for object detection.
\newblock In {\em 2017 IEEE Conference on Computer Vision and Pattern
  Recognition (CVPR)}, pages 936--944, 2017.

\bibitem{lin2017feature}
Tsung-Yi Lin, Piotr Doll{\'a}r, Ross Girshick, Kaiming He, Bharath Hariharan,
  and Serge Belongie.
\newblock Feature pyramid networks for object detection.
\newblock In {\em Proceedings of the IEEE conference on computer vision and
  pattern recognition}, pages 2117--2125, 2017.

\bibitem{manglik2019forecasting}
Aashi Manglik, Xinshuo Weng, Eshed Ohn-Bar, and Kris~M Kitani.
\newblock Forecasting time-to-collision from monocular video: Feasibility,
  dataset, and challenges.
\newblock {\em arXiv preprint arXiv:1903.09102}, 2019.

\bibitem{habitat19iccv}
{Manolis Savva*}, {Abhishek Kadian*}, {Oleksandr Maksymets*}, Yili Zhao, Erik
  Wijmans, Bhavana Jain, Julian Straub, Jia Liu, Vladlen Koltun, Jitendra
  Malik, Devi Parikh, and Dhruv Batra.
\newblock Habitat: {A} {P}latform for {E}mbodied {AI} {R}esearch.
\newblock In {\em Proceedings of the IEEE/CVF International Conference on
  Computer Vision (ICCV)}, 2019.

\bibitem{mcguire2019comparative}
Kimberly~N McGuire, GCHE de Croon, and Karl Tuyls.
\newblock A comparative study of bug algorithms for robot navigation.
\newblock {\em Robotics and Autonomous Systems}, 121:103261, 2019.

\bibitem{mescheder2019occupancy}
Lars Mescheder, Michael Oechsle, Michael Niemeyer, Sebastian Nowozin, and
  Andreas Geiger.
\newblock Occupancy networks: Learning 3d reconstruction in function space.
\newblock In {\em Proceedings of the IEEE Conference on Computer Vision and
  Pattern Recognition}, pages 4460--4470, 2019.

\bibitem{murali2019pyrobot}
Adithyavairavan Murali, Tao Chen, Kalyan~Vasudev Alwala, Dhiraj Gandhi, Lerrel
  Pinto, Saurabh Gupta, and Abhinav Gupta.
\newblock Pyrobot: An open-source robotics framework for research and
  benchmarking.
\newblock {\em arXiv preprint arXiv:1906.08236}, 2019.

\bibitem{park2019deepsdf}
Jeong~Joon Park, Peter Florence, Julian Straub, Richard Newcombe, and Steven
  Lovegrove.
\newblock Deepsdf: Learning continuous signed distance functions for shape
  representation.
\newblock In {\em Proceedings of the IEEE Conference on Computer Vision and
  Pattern Recognition}, pages 165--174, 2019.

\bibitem{pinto2016supersizing}
Lerrel Pinto and Abhinav Gupta.
\newblock Supersizing self-supervision: Learning to grasp from 50k tries and
  700 robot hours.
\newblock In {\em 2016 IEEE international conference on robotics and automation
  (ICRA)}, pages 3406--3413. IEEE, 2016.

\bibitem{Purushwalkam19}
Senthil Purushwalkam, Abhinav Gupta, Danny~M. Kaufman, and Bryan~C. Russell.
\newblock Bounce and learn: Modeling scene dynamics with real-world bounces.
\newblock {\em CoRR}, abs/1904.06827, 2019.

\bibitem{ramakrishnan2020occupancy}
Santhosh~K Ramakrishnan, Ziad Al-Halah, and Kristen Grauman.
\newblock Occupancy anticipation for efficient exploration and navigation.
\newblock {\em ECCV 2020}, 2020.

\bibitem{ILSVRC15}
Olga Russakovsky, Jia Deng, Hao Su, Jonathan Krause, Sanjeev Satheesh, Sean Ma,
  Zhiheng Huang, Andrej Karpathy, Aditya Khosla, Michael Bernstein,
  Alexander~C. Berg, and Li Fei-Fei.
\newblock {ImageNet Large Scale Visual Recognition Challenge}.
\newblock {\em International Journal of Computer Vision (IJCV)},
  115(3):211--252, 2015.

\bibitem{pifuSHNMKL19}
Shunsuke Saito, , Zeng Huang, Ryota Natsume, Shigeo Morishima, Angjoo Kanazawa,
  and Hao Li.
\newblock Pifu: Pixel-aligned implicit function for high-resolution clothed
  human digitization.
\newblock {\em arXiv preprint arXiv:1905.05172}, 2019.

\bibitem{saito2019pifu}
Shunsuke Saito, Zeng Huang, Ryota Natsume, Shigeo Morishima, Angjoo Kanazawa,
  and Hao Li.
\newblock Pifu: Pixel-aligned implicit function for high-resolution clothed
  human digitization.
\newblock In {\em Proceedings of the IEEE International Conference on Computer
  Vision}, pages 2304--2314, 2019.

\bibitem{Scharstein02}
Daniel Scharstein and Richard Szeliski.
\newblock A taxonomy and evaluation of dense two-frame stereo correspondence
  algorithms.
\newblock {\em IJCV}, 47(1):7--42, 2002.

\bibitem{shrestha2019learned}
Rakesh Shrestha, Fei-Peng Tian, Wei Feng, Ping Tan, and Richard Vaughan.
\newblock Learned map prediction for enhanced mobile robot exploration.
\newblock In {\em 2019 International Conference on Robotics and Automation
  (ICRA)}, pages 1197--1204. IEEE, 2019.

\bibitem{Smith2005}
Linda Smith and Michael Gasser.
\newblock The development of embodied cognition: Six lessons from babies.
\newblock {\em Artif. Life}, 11(1–2):13–30, Jan. 2005.

\bibitem{thrun2002probabilistic}
Sebastian Thrun.
\newblock Probabilistic robotics.
\newblock {\em Communications of the ACM}, 45(3):52--57, 2002.

\bibitem{drcTulsiani17}
Shubham Tulsiani, Tinghui Zhou, Alexei~A. Efros, and Jitendra Malik.
\newblock Multi-view supervision for single-view reconstruction via
  differentiable ray consistency.
\newblock In {\em Computer Vision and Pattern Recognition (CVPR)}, 2017.

\bibitem{tung2017self}
Hsiao-Yu~Fish Tung, Hsiao-Wei Tung, Ersin Yumer, and Katerina Fragkiadaki.
\newblock Self-supervised learning of motion capture.
\newblock {\em arXiv preprint arXiv:1712.01337}, 2017.

\bibitem{watson2019self}
Jamie Watson, Michael Firman, Gabriel~J Brostow, and Daniyar Turmukhambetov.
\newblock Self-supervised monocular depth hints.
\newblock In {\em Proceedings of the IEEE/CVF International Conference on
  Computer Vision}, pages 2162--2171, 2019.

\bibitem{xiazamirhe2018gibsonenv}
Fei Xia, Amir R.~Zamir, Zhi-Yang He, Alexander Sax, Jitendra Malik, and Silvio
  Savarese.
\newblock Gibson env: real-world perception for embodied agents.
\newblock In {\em Computer Vision and Pattern Recognition (CVPR), 2018 IEEE
  Conference on}. IEEE, 2018.

\bibitem{xu2019disn}
Qiangeng Xu, Weiyue Wang, Duygu Ceylan, Radomir Mech, and Ulrich Neumann.
\newblock Disn: Deep implicit surface network for high-quality single-view 3d
  reconstruction.
\newblock In {\em Advances in Neural Information Processing Systems}, pages
  492--502, 2019.

\bibitem{yang2018visual}
Wei Yang, Xiaolong Wang, Ali Farhadi, Abhinav Gupta, and Roozbeh Mottaghi.
\newblock Visual semantic navigation using scene priors.
\newblock {\em arXiv preprint arXiv:1810.06543}, 2018.

\bibitem{zhou2017unsupervised}
Tinghui Zhou, Matthew Brown, Noah Snavely, and David~G. Lowe.
\newblock Unsupervised learning of depth and ego-motion from video.
\newblock In {\em CVPR}, 2017.

\bibitem{zou2018layoutnet}
Chuhang Zou, Alex Colburn, Qi Shan, and Derek Hoiem.
\newblock Layoutnet: Reconstructing the 3d room layout from a single rgb image.
\newblock In {\em Proceedings of the IEEE Conference on Computer Vision and
  Pattern Recognition}, pages 2051--2059, 2018.

\end{thebibliography}
}

\clearpage
\setcounter{section}{0}
\renewcommand{\thesection}{\Alph{section}}
\section{Remote Prediction Model Details}
\label{sec:remotemodel_supp}

All remote point prediction models follow a PIFu \cite{saito2019pifu} style architecture composed of two components:
\begin{enumerate}
\item an {\it Image Encoder} which predicts features used to condition implicit functions at each pixel.
\item an {\it Implicit Function Decoder} that maps from the feature map (plus auxiliary information) to an output that is target-dependent.
\end{enumerate}

\par \noindent
{\bf Image Encoder:} All remote point prediction models use the same image encoder, a feature-pyramid network \cite{fpn} applied to a ResNet-18 CNN \cite{He2015} pretrained on ImageNet \cite{ILSVRC15}. Given a $256 \times 256 \times 3$ input image, the encoder produces a pyramid of features with feature dimensions of 64, 64, 128, 256 and 512. Each of these features can be indexed with a (possibly fractional) coordinate corresponding the projected location of each query point. Fractional coordinates are accessed using bi-linear interpolation. The indexed features are then concatenated into a 1024-dimensional feature vector to condition the implicit function for each query point. 

\vspace{2mm}
\par \noindent
{\bf Implicit Function Decoder:} 
The implicit function decoder maps a feature vector sampled at a location to an output. 

\vspace{2mm}
\par \noindent {\it Input:} 
The decoder takes as input a feature vector that is bi-linearly sampled from the image encoder as described above. In all cases, we concatenate a scalar to this representing the projected depth $d_\pi(\pointx)$, forming the feature denoted $\phi(\pointx)$ in the main paper. Methods that use the relative heading angle as an input also have information about the angle concatenated to the feature; we encode angles using a sine/cosine encoding of $[\sin(\theta), \cos(\theta)]$. Thus angle-agnostic methods have an input dimension of $1025$ and angle-conditioned ones have an input dimension of $1027$.

\vspace{2mm}
\par \noindent{\it Network:}
We decode the above concatenated feature vector with a multi-layer perceptron with hidden layer sizes 1024, 512, and 256; each layer is followed by a Leaky ReLU non-linearity (with negative slope 0.01). 

\vspace{2mm}
\par \noindent{\it Output and Loss Function:} 
The output size and loss function used depends on the model:
\begin{itemize}
    \item {\it Classification:} In the main multi-class case, the network produces a 11-dimensional output corresponding to the logits for steps $(0, 1, \ldots, 10+)$. The resulting loss function is the cross-entropy between the softmax of these logits and a target value derived from collision replay.
    \item {\it Regression:} In the regression case, the network produces a 1-dimensional output corresponding to the log of the number of steps. The loss function is the Smoothed L1, or L2, loss between this and the log of the target number of steps. 
    \item {\it Binary Freespace:} In comparisons with freespace classification, we produce a 1-dimensional output representing the log-odds that the space is free. The loss function is the binary cross entropy between this and the target value.
\end{itemize}

\section{Distance Function Decoding Details}
\label{sec:decoding_supp}

Once the networks have made predictions, we need to decode these into distance functions. 

\vspace{2mm}
\par \noindent
{\bf Classification Minima Distance Function Decoding:}
As described in Section 3.2 in the main paper, when computing a distance function in the multi-class classification case, we find the minimum step value where the cumulative probability of collision exceeds a threshold $\epsilon$ (i.e. $(\sum_{j \le t} P(j)) \ge \epsilon$). Applied na\"ively, this leads to discrete jumps between bins and always over-estimates the time to collision. We therefore linearly interpolate between the bins. This reduces over-prediction and leads to smoothly varying distance function predictions.
\vspace{2mm}
\par \noindent
{\bf Angle Minima Distance Function Decoding:}
In experiments considering the heading angle of each remote point, we must take the minimum over possible angles to produce a distance function for the scene. In the remote prediction case, we evaluate this by taking the minimum over 32 evenly spaced sample angles between $0^\circ$ and $360^\circ$. In the egocentric case, we evaluate the minimum over the actions.

\section{Egocentric Prediction Model Details}
\label{sec:supp_egocentric}

\par\noindent{\bf Backbone:} In the egocentric setting, we apply a ResNet-18 backbone to each input image, followed by average pooling and then a multi-layer perceptron with hidden layer sizes of 512 and 256, using the same Leaky ReLU activation with a negative slope of 0.01. We then compute an output layer, with shape depending on whether we use classification or regression as an objective following 

\begin{itemize}
    \item {\it Classification:} In the classification case, we produce an output of shape $11 \times 3$, which we interpret as a logits for the conditional probability distribution $P(t|\alpha)$, where t is one of the discrete values $(0, 1, \ldots, 10+)$, and $\alpha$ is one of the three actions ({\it turn left}, {\it move forwards}, {\it turn right}). We do not consider the {\it turn around} action for egocentric training, as there is reduced information in an egocentric view to predict whether the space behind the agent is occupied. At train time, we take examples of the form (image I, action A, steps to collision T), and apply a Cross Entropy loss between $P(t|\alpha=A)$ and the label T.
    \item {\it Regression:} In the regression model, we mirror the classification case as closely as possible by creating a 3 dimensional output vector, where each element predicts a steps to collision value conditioned on the agent taking a particular action. At train time, we supervise the value from this vector corresponding to the action taken in a particular example. We use the same Smoothed L1, or L2, regression objective as in the remote prediction case. 
\end{itemize}

\section{Data Collection Details}
\label{sec:data_supp}

We filter the points and images that are used for training. For start with a dataset of 500 steps for each of 10 episodes from 360 different environments (for our main Gibson dataset), for a total of 1.8M total images. We discard any training images in this dataset containing fewer than 1 collision point or fewer than 5 non-collision points within the camera frustum. This yields a dataset of 800K images.

\section{Training Details}
\label{sec:training_supp}

\vspace{2mm}
\par \noindent {\bf Optimization:} At train time, we use a batch size of 128 images. In the remote point prediction setting, we randomly sample 150 trajectory points to query in each example. We train each remote prediction model for 20 epochs over the above 800K image dataset. We train egocentric models for 5 epochs, which maximizes performance on the validation set. All models are optimized with Adam \cite{Kingma2015}, following a cosine schedule with a starting value of 2e-5, a maximum value of 2e-4 after 30\% of training has elapsed, and a final value of 1e-5. 
Training a remote prediction model takes approximately 12 hours on two Nvidia 2080Ti GPUs.

\vspace{2mm}
\par \noindent {\bf Data Augmentation:} We apply augmentation in the form of random horizontal flips, and random horizontal shifts with percentage shift distributed normally with $\sigma = 10\%$. All image augmentations are equivalently applied to the camera matrices, so that trajectory points still correctly line up with image contents. In the remote prediction setting, we also apply 2D Gaussian noise with $\sigma = 3cm$ within the floor plane to the query points' scene coordinates, to encourage smoothness and maximize the number of pixels being supervised across epochs.
\section {Comparison to Learning to Fly by Crashing}
\label{sec:lfc_supp}

Learning to Fly By Crashing (LFC) \cite{gandhi2017learning} is a prior work which uses real world drone trajectories to learn a model to predict whether a drone's egocentric image is 'near' or 'far' from colliding with the environment, which approximately translates to predicting whether the agent is within \textit{K} steps of colliding, for \textit{K} chosen at training time. 

To enable comparison, we created versions of our model which emulate LFC's original binary training task. LFC is most similar to egocentric setting, so we adopt the same prediction backbone and training methods as is described in Section \ref{sec:supp_egocentric}. We replace the usual regression or multi-class classification output with a single scalar output. We then use a binary cross entropy loss, with labels specifying whether the steps-to-collision value is greater or less than a given K. We trained a version of this LFC-analogous model for K = 2, 4, 6, 8 in order to cover the 0 to 10 step range modelled by our main method.

Distance function are  difficult  since  LFC  is  aimed  at  producing  a  policy  as opposed to scene structure or a distance function. Nonetheless, we compare on both our own task of producing a scene distance function, as well as LFC's original training task of distinguishing whether the given image is within K steps of colliding. To evaluate LFC on our distance function task we linearly re-scale it's output probability up to the same 0-4m range as our main method. To evaluate our multi-class classification model on LFC's task we produce a binary value by determining if the most likely step count (argmax) is greater or less than K. %

In Table \ref{tab:lfc_quantitative_dists} we observe that our method outperforms LFC in distance function prediction across all K as expected. Additionally, as seen in Table \ref{tab:lfc_quantitative_binary} our method outperforms LFC on its binary classification task, likely due to multi-task learning effects.

\begin{table}[]
    \centering
    \caption{Quantitative results for distance function decoding}
    \label{tab:egocentric}
    \begin{tabular}{lccc} \toprule
        & \multicolumn{3}{c}{Distance Function} 
        \\
        & MAE & RMSE & $\% \le t$ 
        \\ \midrule
        LFC (K=2) & 1.30 & 1.46 & 0.101
        \\
        LFC (K=4) & 0.95 & 1.17 & 0.22
        \\
        LFC (K=6) & 0.73 & 0.96 & 0.294
        \\ 
        LFC (K=8) & 0.61 & 0.841 & 0.335
        \\
        Regression & 0.96 & 1.14 & 0.13
        \\
        Classification & \textbf{0.58} & \textbf{0.79} & \textbf{0.36}
        \\
        \bottomrule
    \end{tabular}
    
    \label{tab:ego_quantitative}
    \label{tab:lfc_quantitative_dists}
\end{table}

\begin{table}[]
    \centering
    \caption{Quantitative results for threshold binary classification}
    \label{tab:egocentric}
    \begin{tabular}{lcccc} \toprule
        & \multicolumn{4}{c}{F1 Score} 
        \\
        & K=2 & K=4 & K=6 & K=8
        \\ \midrule
        LFC & 0.85 & 0.77 & 0.72 & 0.62
        \\
        Ours & \textbf{0.86} & \textbf{0.79} & \textbf{0.74} & \textbf{0.66}
        \\
        \bottomrule
    \end{tabular}
    
    \label{tab:lfc_quantitative_binary}
\end{table}
\section{Theoretical Modeling}
\label{sec:theoretical_supp}

\begin{figure*}[t]
    \centering
    \begin{tabular}{@{~}c@{~}c@{~}c@{~}c@{~}}
    \includegraphics[width=0.24\linewidth]{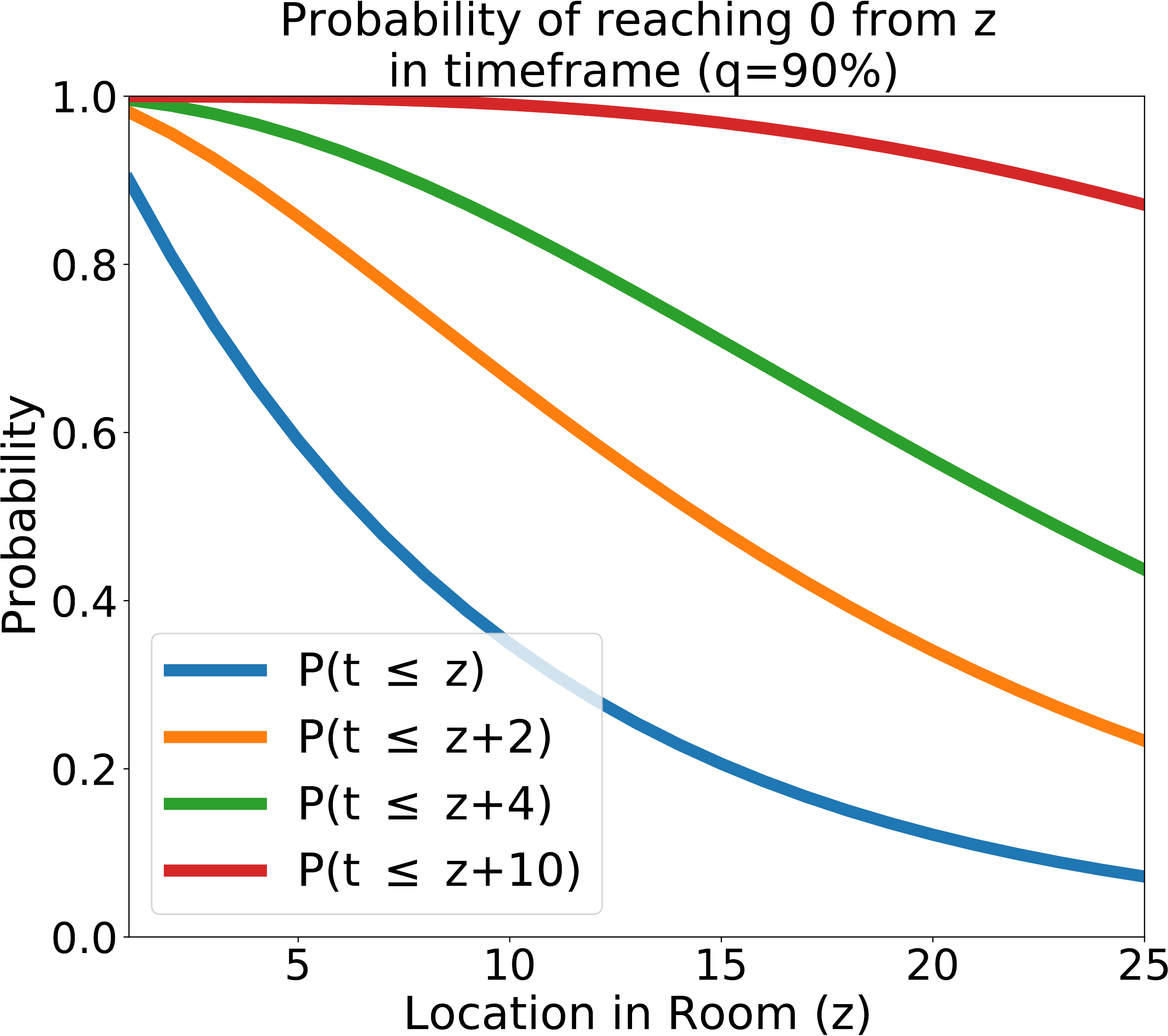} &
    \includegraphics[width=0.24\linewidth]{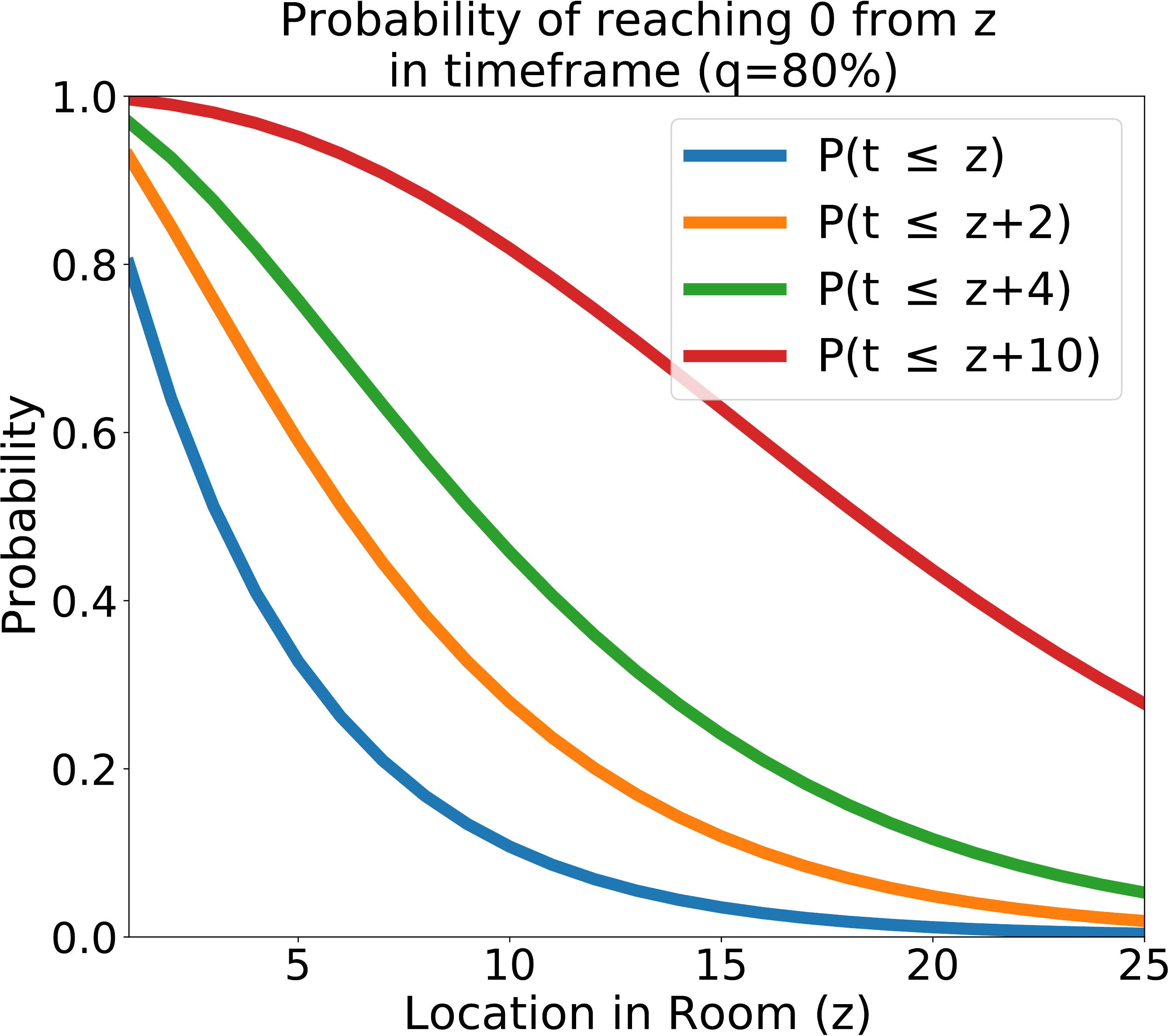} &    
    \includegraphics[width=0.24\linewidth]{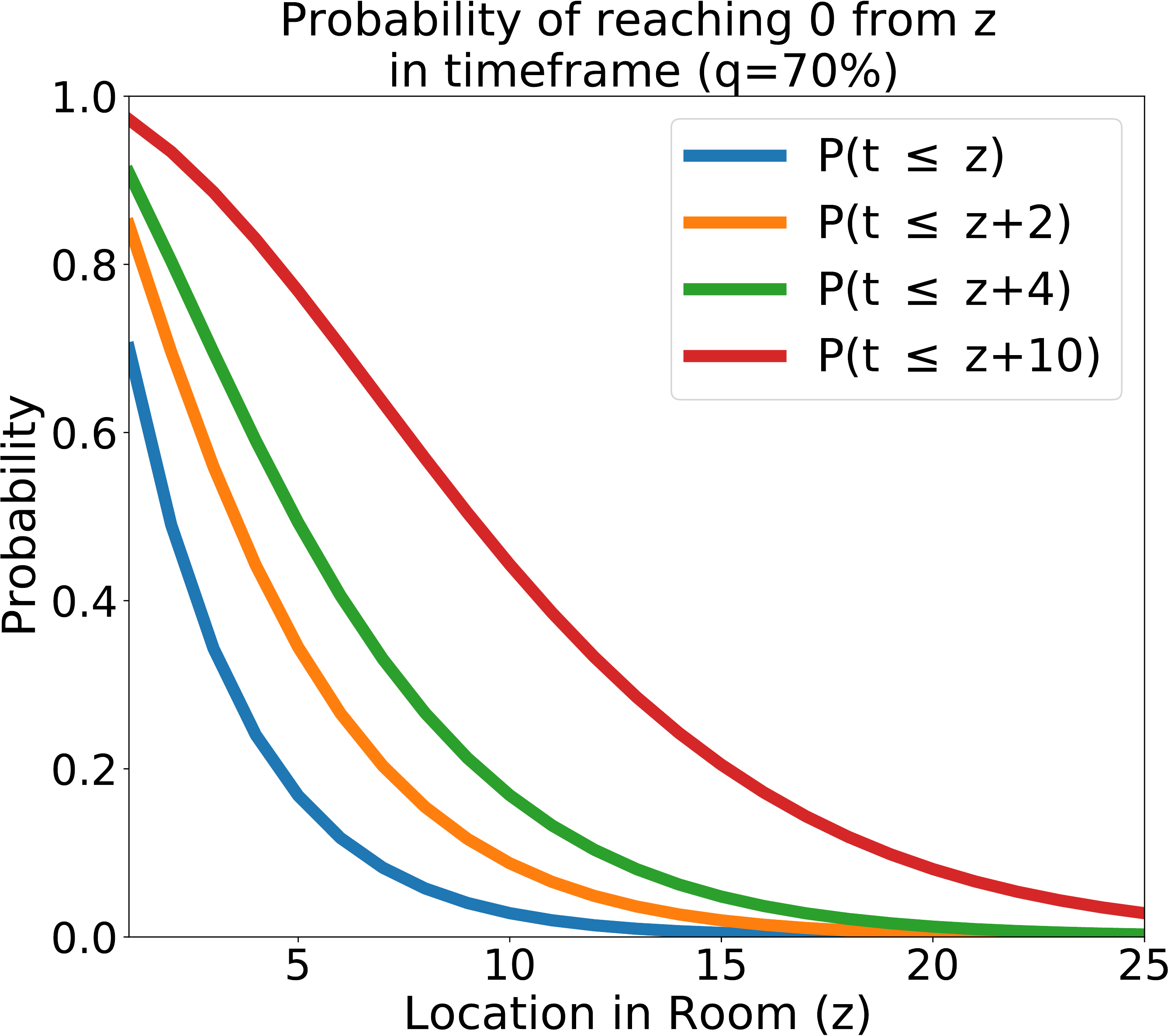} &    
    \includegraphics[width=0.24\linewidth]{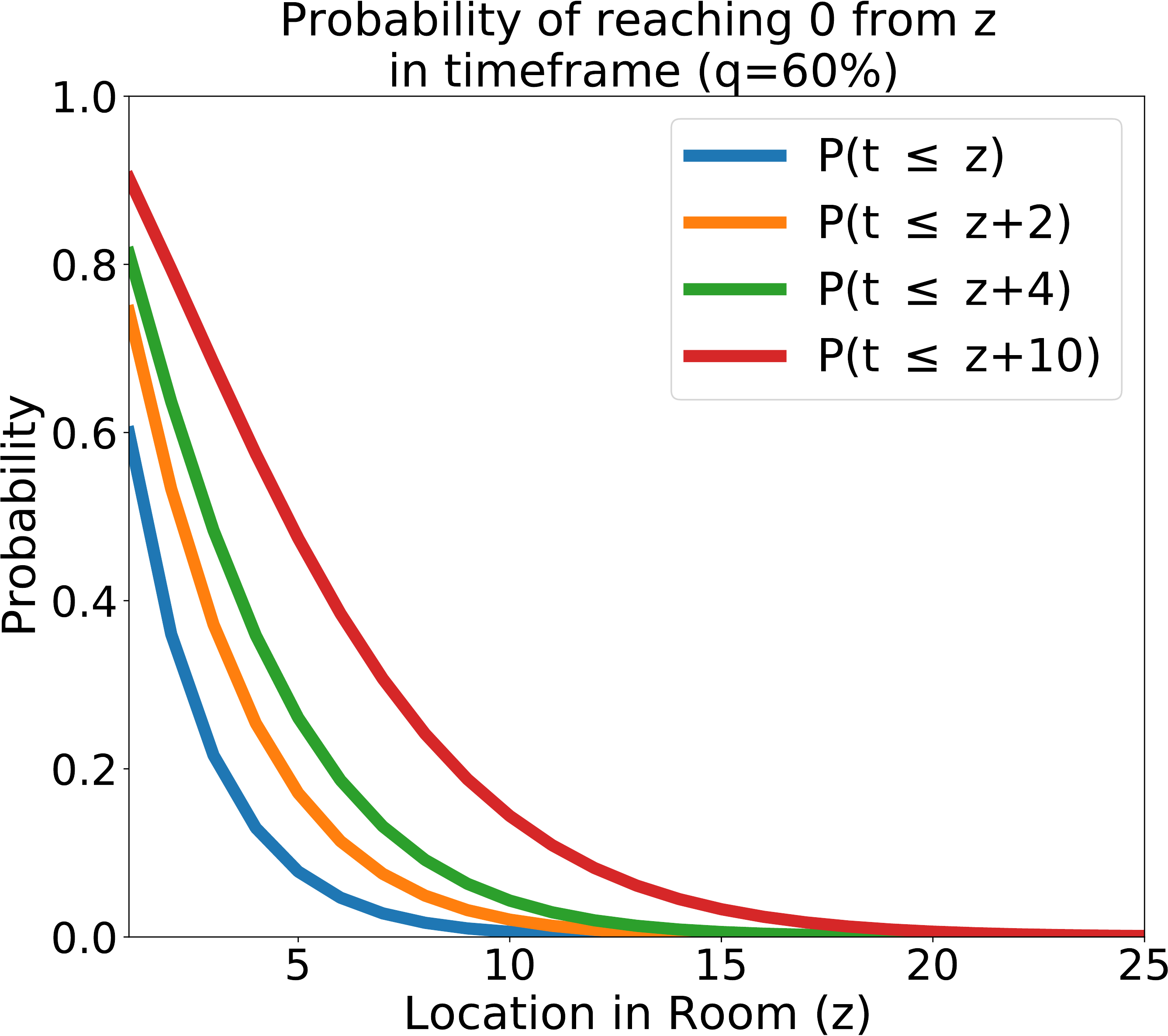} \\       
    \includegraphics[width=0.24\linewidth]{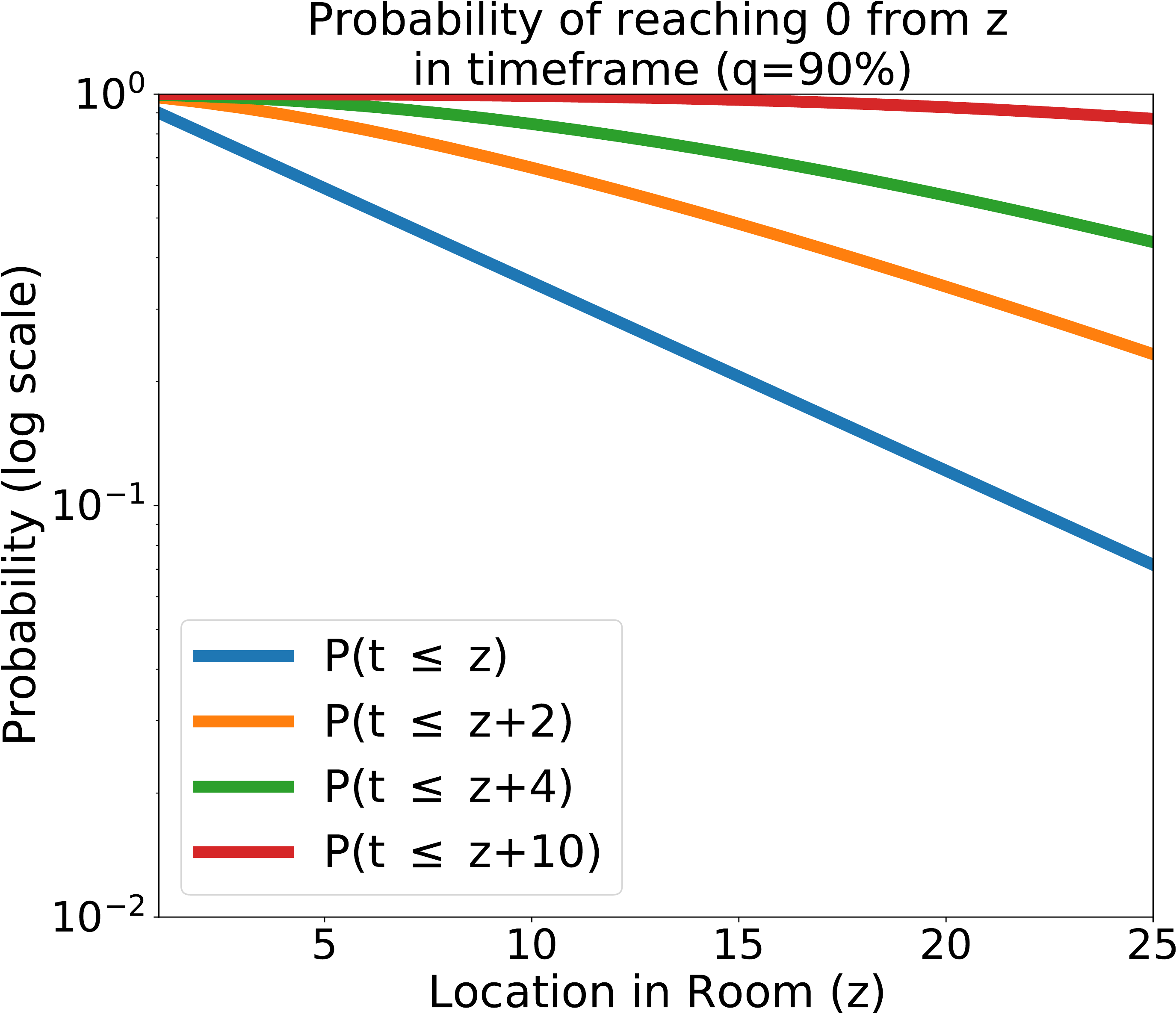} &
    \includegraphics[width=0.24\linewidth]{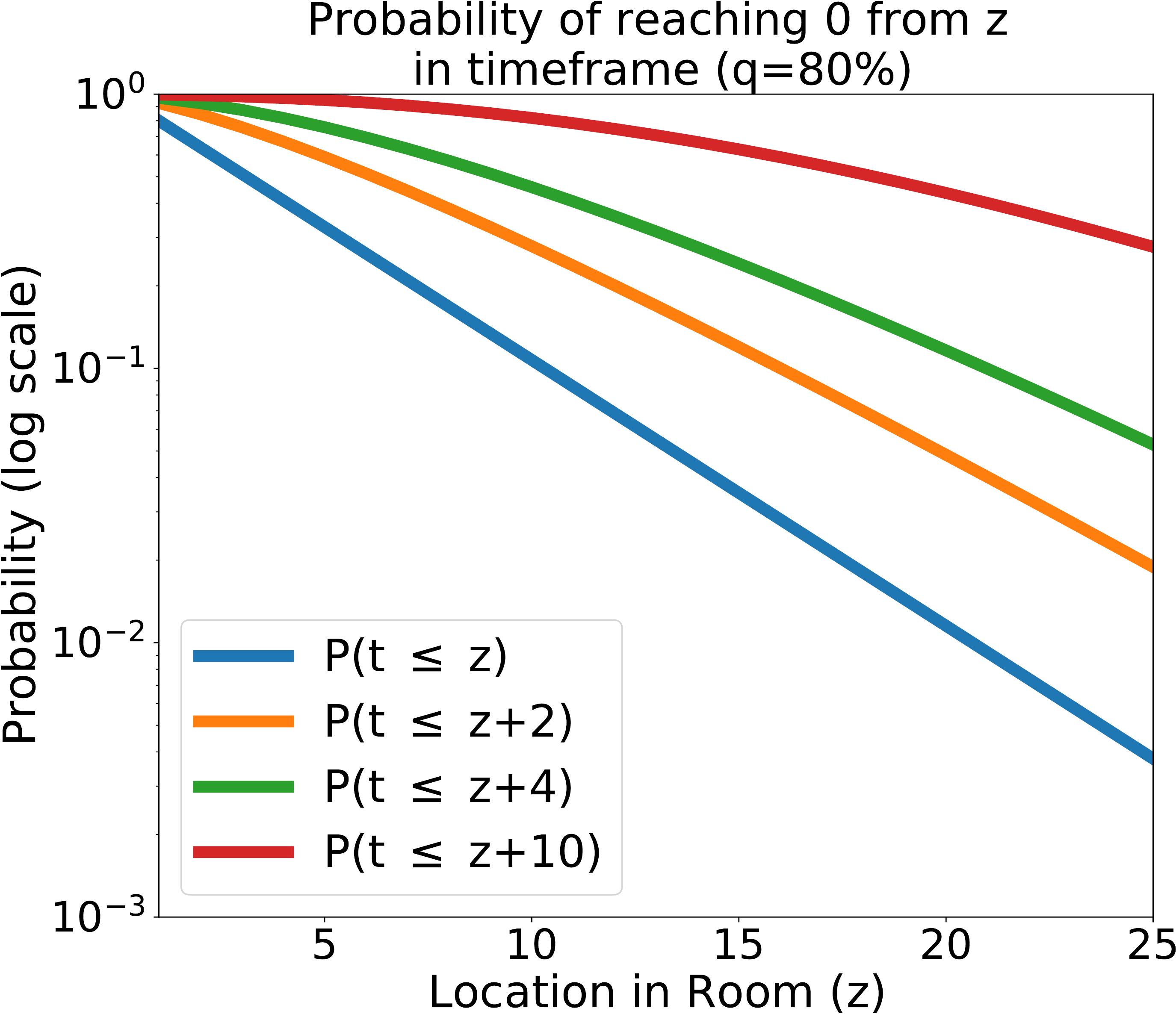} &    
    \includegraphics[width=0.24\linewidth]{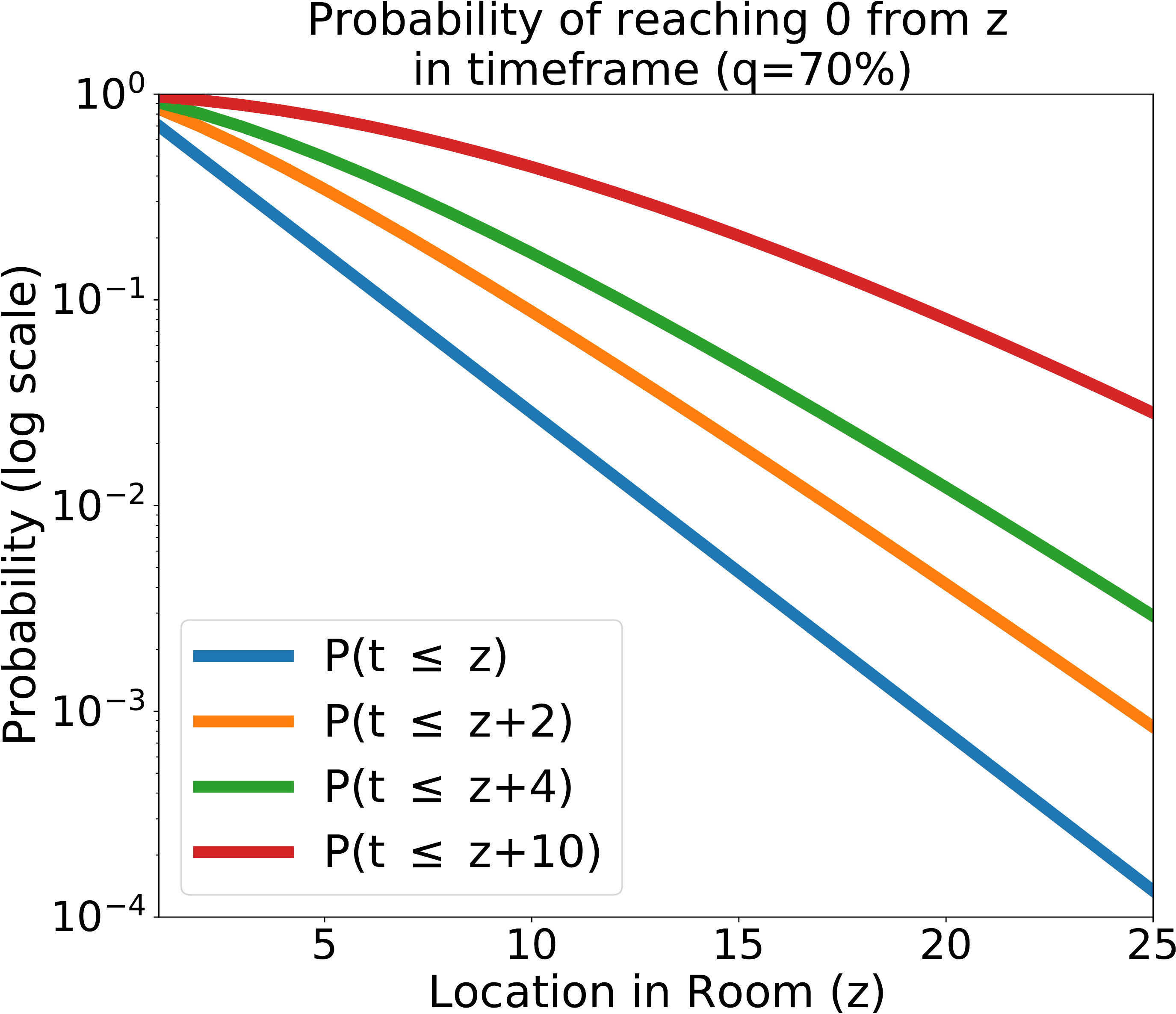} &
    \includegraphics[width=0.24\linewidth]{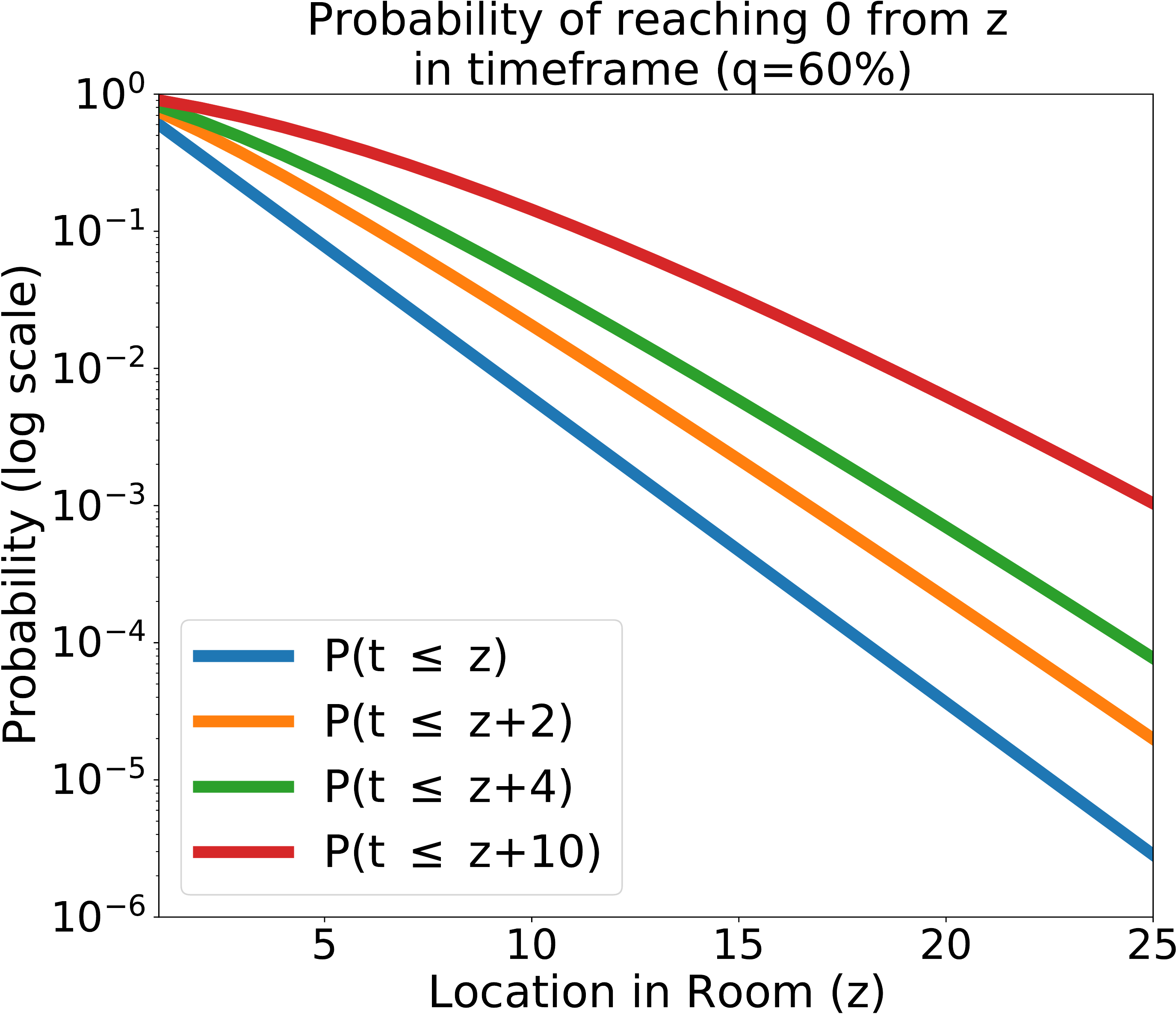} \\
    \end{tabular}
    \caption{Plots of the probability of seeing a fairly short path as a function of location inside the room. Each figure plots probability of taking the shortest path {\bf within} a set of tolerances as a function of the location in the room. In other words, the $P(t \le z+4)$ plot
    is the probability of having a path that is no more than $4$ steps farther than the optimal path and is calculated via $\sum_{i=1}^{z+4} P_T(t=i)$. The columns vary the probability of moving left $q$ and the rows show linear (top) and logarithmic (bottom) scales. 
    If the agent has a reasonably high chance of moving left ($q=90\%$), then exactly short paths are surprisingly common. Even with a lower chance of moving left, nearly shortest paths are fairly common and well-represented in the training data.}
    \label{fig:distributions}
\end{figure*}

\begin{figure}[t]
    \centering
    \begin{tabular}{@{~}c@{~}c@{~}}
    \includegraphics[width=0.475\linewidth]{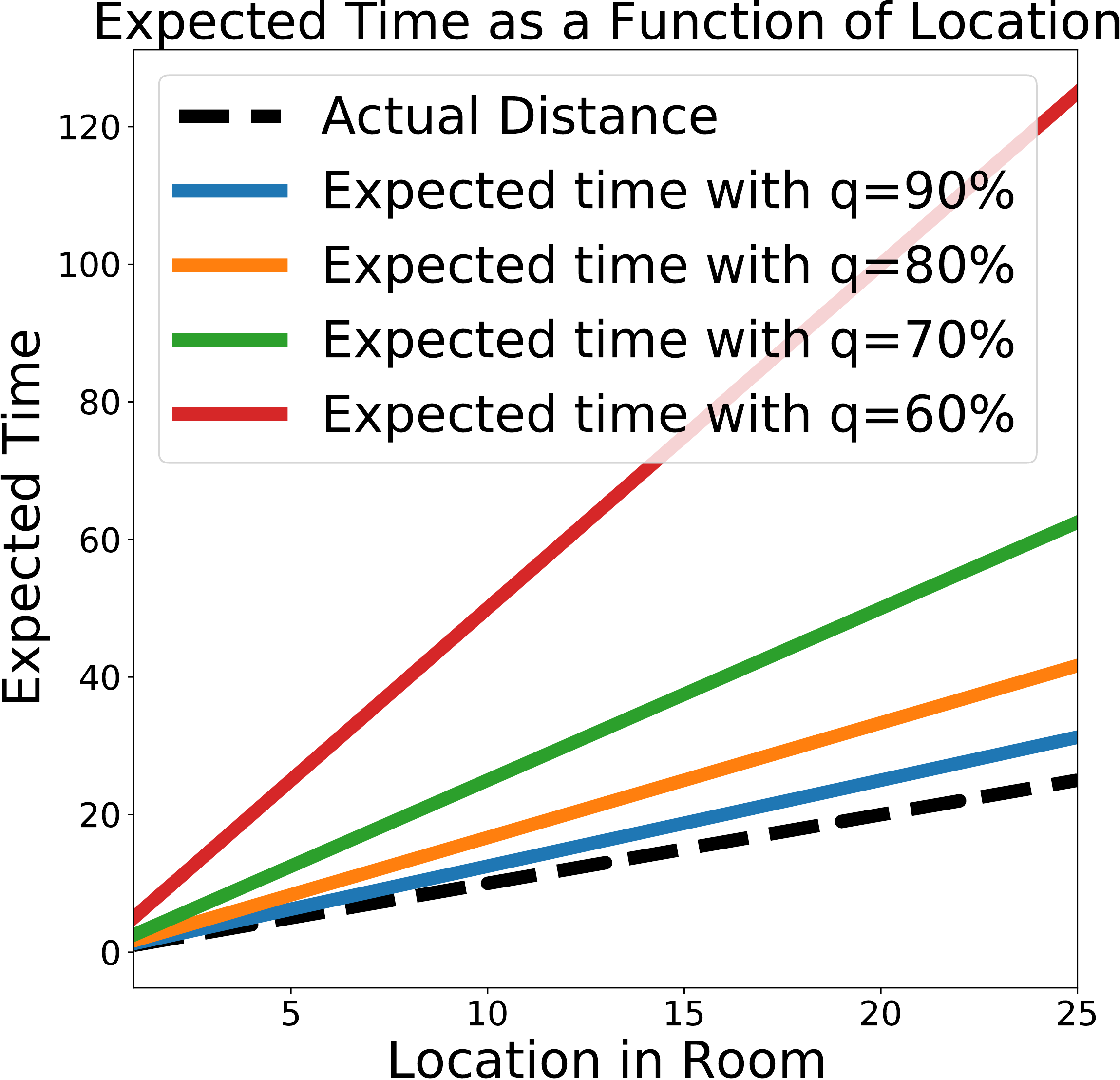} &
    \includegraphics[width=0.475\linewidth]{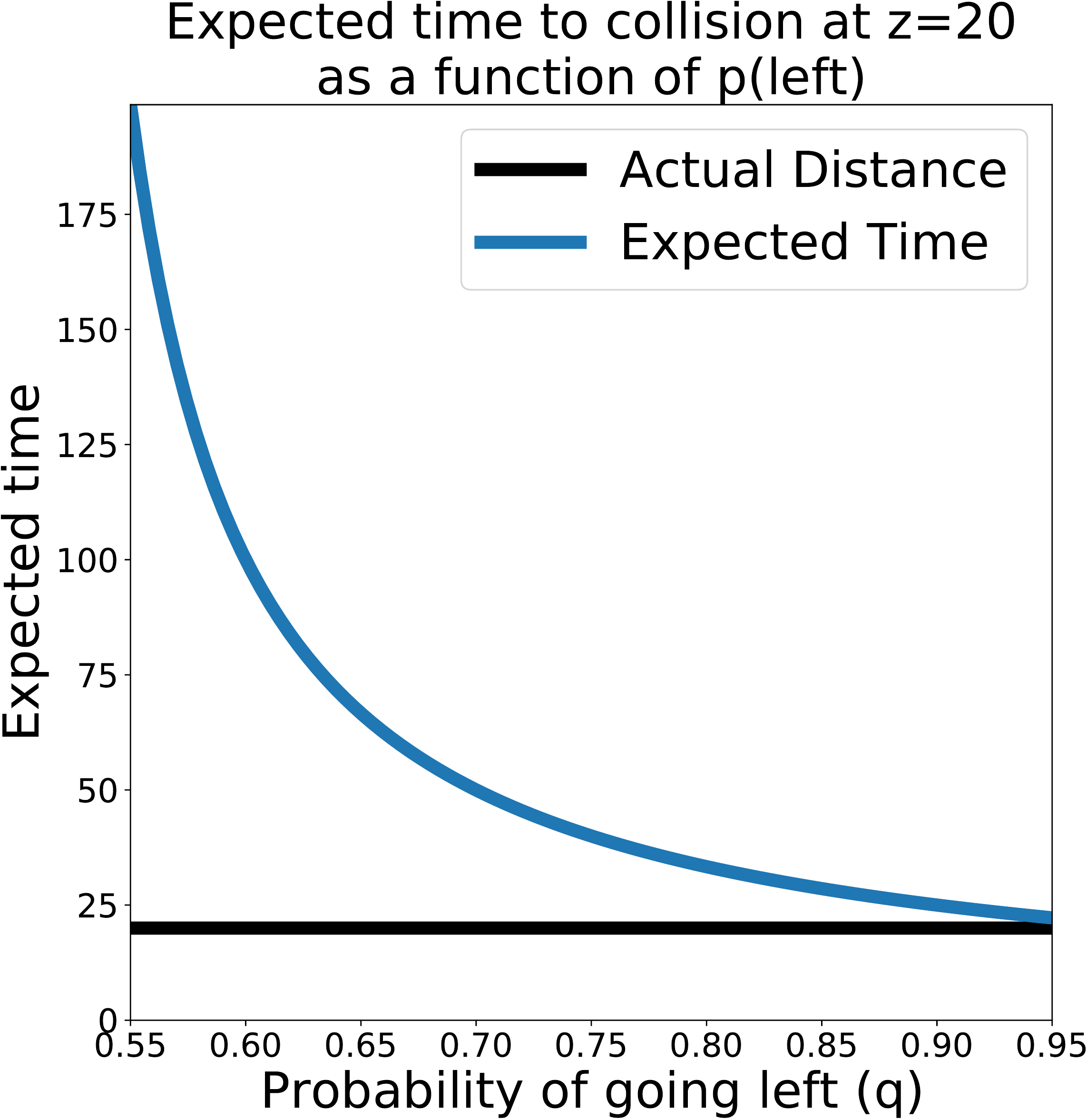} \\
    (a) & (b) \\
    \end{tabular}
    \caption{Expected time: (a) as a function of location for a set of $q$; (b) as a function of $q$ for a fixed location.}
    \label{fig:expected}
\end{figure}

Our setting can be connected with classic random walk problem settings such as the Gambler's ruin. These settings, albeit in simplified worlds, enable us to derive analytical results describing times to collisions. In turn, these analytical results explain empirical behavior of our system and of baselines. More specifically, if can characterize how likely particular path lengths are, we can
ask questions like: is it likely that we will see short paths during training? what path length do we expect?

\vspace{2mm}
\par \noindent {\bf Setting:} Suppose an agent is in a 1D grid world where cells $1,\ldots, a-1$ are free and there are walls
at cell $0$ and $a$ that trigger the collision sensors. The agent starts at a 
position $z$ and moves to the right with probability
$p$ and to the left with probability $q=1-p$. This corresponds precisely to the Gambler's ruin (our results, and thus notation, follow Feller \cite{Feller50}, Chapter 14), where a gambler and casino repeatedly play a fixed game where
the gambler wins dollar with probability $p$ and the casino wins a dollar with probability $q$. The gambler starts with $z$ dollars (an integer) and the casino with $a-z$ dollars and both play until either the gambler has $0$ dollars ({\it ruin}) or wins $a$ dollars ({\it gain}). Gambler's ruin problems focus on characterizing the number of rounds the gambler and casino play. While well studied in the stochastic processes area and readily simulated, closed form solutions to these problems resist casual solutions since the number of rounds is unknown in advance.

There are a number of results that are of interest to our setting. All but the shortest
path one appear in Feller \cite{Feller50}. We will introduce them and then show how these are of use. Suppose $T$ is the random variable representing the time to collision whose distribution depends on ($p$, $a$, and $z$). Given that our agents act with some direction, we are interested in games that are biased (or $p \not = q$). We will model this with $p < \frac{1}{2}$. This means the gambler is likely to have ruin or, in our setting, the agent is likely to hit the left wall; one can reverse the roles of gambler and casino to obtain results for hitting the right wall. We focus on the case of ruin (i.e., hitting the left wall). The most most simple results are given for time to ruin, and ruin serves as a reasonable proxy for time to collision because ruin is virtually guaranteed for $p < \frac{1}{2}$ and reasonably-sized $a$).

\vspace{2mm}
\noindent {\bf Probability of Ruin:} We are generally concerned with settings where the agent moves forward with high probability. For a fair ($p = q$) game, the probability of hitting the left wall/ruin is $1-z/a$. For an unfair game ($p \not = q$), the probability is given by \cite{Feller50} as
\begin{equation}
    \frac{(q/p)^a - (q/p)^z}{(q-p)^a -1},
\end{equation}
which becomes extraordinarily likely high as $z$ and $a$ get bigger so long as there is some advantage for the house (i.e., the agent is more likely to go left than right).

\vspace{2mm}
\noindent {\bf Expected duration:} the expected duration of the game (i.e., $E[T]$) is important
because the expected value of a distribution is the minimum for predicting samples for that distribution. Thus, a system that models time to collisions by minimizing the L2 error 
between samples from that distribution and the prediction has a minimum at the
expected value of the distribution. The expected value has a 
clean, closed form expression of $z(a-z)$ for $p=\frac{1}{2}$ and 
\begin{equation}
E[T] = \frac{z}{q-p} - \frac{a}{q-p} \cdot \frac{1 - (q/p)^z}{1-(q/p)^a}
\end{equation}
for $p \not = \frac{1}{2}$. In Fig.~\ref{fig:expected}, we plot some plots of the expected time as (a) a function of location for a set of $q$s; and (b) a function of $q$ for a fixed location. 

This $E[T]$ is important because a network trained to minimize the $\textrm{L}_2$ distance / MSE between its predictions and samples from $T$ has its minimum on at $E[T]$. Across all locations and probabilities of moving towards the nearest wall $q$, $E[T] > z$ and is an overestimate by a factor that depends on $q$.

\vspace{2mm}
\noindent {\bf Probability of seeing the shortest path to the wall:} One might wonder
how frequently we would sample a {\it shortest} path to the wall given that we observe a time to collision. Without loss of generality, assume that $z \le a/2$: the shortest path to any wall goes to the left and takes $z$ steps. The probability of seeing the shortest path given a sample is
then given by $q^z$. This can be seen by noting that one only has to look at the tree of possible
paths  after $z$ steps have played out. If all steps go to the
left, then there is a collision; otherwise the result is not a shortest path. The probability
of seeing the shortest path is relatively small for large rooms.

\vspace{2mm}
\noindent {\bf Distribution over probability of ruin in $t$ steps.}  If the shortest path takes $t=z$ steps, we may be equally happy to 
reach the nearest surface in $t=z+2$ or $t=z+4$ steps since these distinctions may
be washed out in actuation noise. There are known results for the particular case where we
are only concerned with time to arrival at the leftmost wall or ruin. This helps
us characterize how frequently we might see nearly-optimal paths
to the wall.
This is given by the involved but analytical formula \cite{Feller50} (replacing Feller's $n$ with our $t$)
\begin{equation}
\begin{split}
p_T(t|\textrm{ruin}) = & a^{-1} 2^t p^{(t-z)/2} q^{(t+z)/2}\\
       & \sum_{v=1}^{a-1} 
\cos^{t-1}\left(\frac{\pi v}{a}\right) 
\sin\left(\frac{\pi v}{a}\right)
\sin\left(\frac{\pi z v}{a} \right),
\end{split}
\end{equation}
which gives (assuming ruin occurs), the probability of it occurring on the $t$th step.
This is a $p_T(t|ruin)$ rather than $p_T(t)$, the overwhelming likelihood of 
ruin means this gives close agreement to empirical simulation results for termination.
This underestimates probabilities of getting a short path in the middle of rooms
but otherwise gives good agreement.

We plot distributions of $P_T$ in Fig.~\ref{fig:distributions} for a large room $a = 51$: note that
with a step size of $25$cm, this room would be $12.5$m across. The figure shows
that if the agent has a reasonably high chance of moving towards the nearest wall ($q=90\%$), then exactly short paths are surprisingly common. Even with a lower chance of moving towards the wall, nearly shortest paths are fairly common and well-represented in the training data. The only conditions under which one is unlikely to see fairly short paths is when the agent wanders ($q$ small) or the room is enormous ($a$ very large). Wandering can be fixed by picking a more forward-moving policy; the room can be fixed by increasing the step size, which effectively reduces $a$.

\section{Qualitative Results Tables}
\label{sec:qualitative_supp}

\begin{figure*}
\centering\includegraphics[scale=0.7]{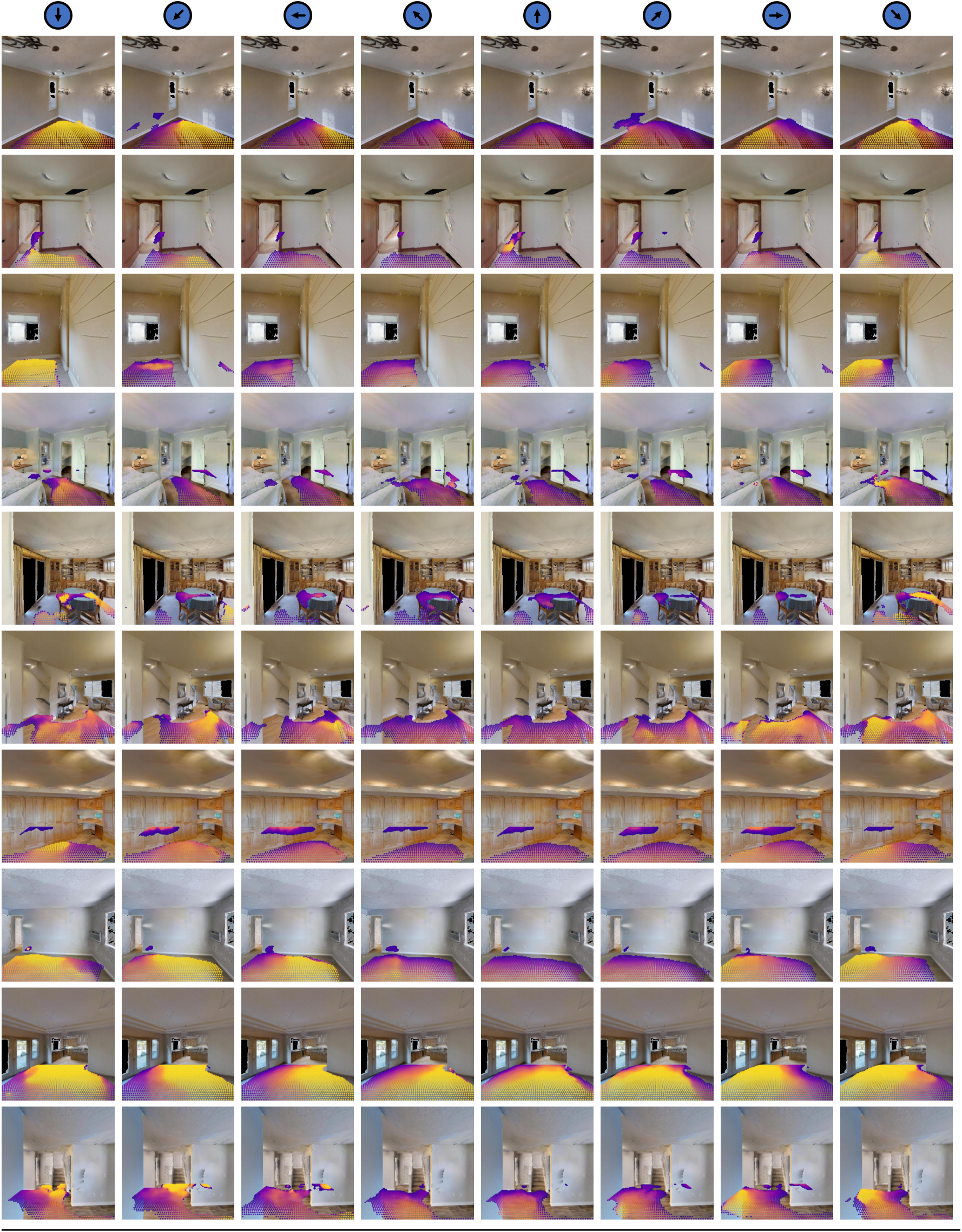}
\caption{Randomly selected examples of the conditional probability $P(t | \alpha)$ for grids of scene points and a varying heading angle $\alpha$ as shown by markers in blue. Probabilities are obtained from a classification model trained with noisy random walks.}
\label{fig:angles_supplement}
\end{figure*}

\begin{figure*}
    \centering
    \begin{minipage}{.3\textwidth}
        \includegraphics[width=\linewidth]{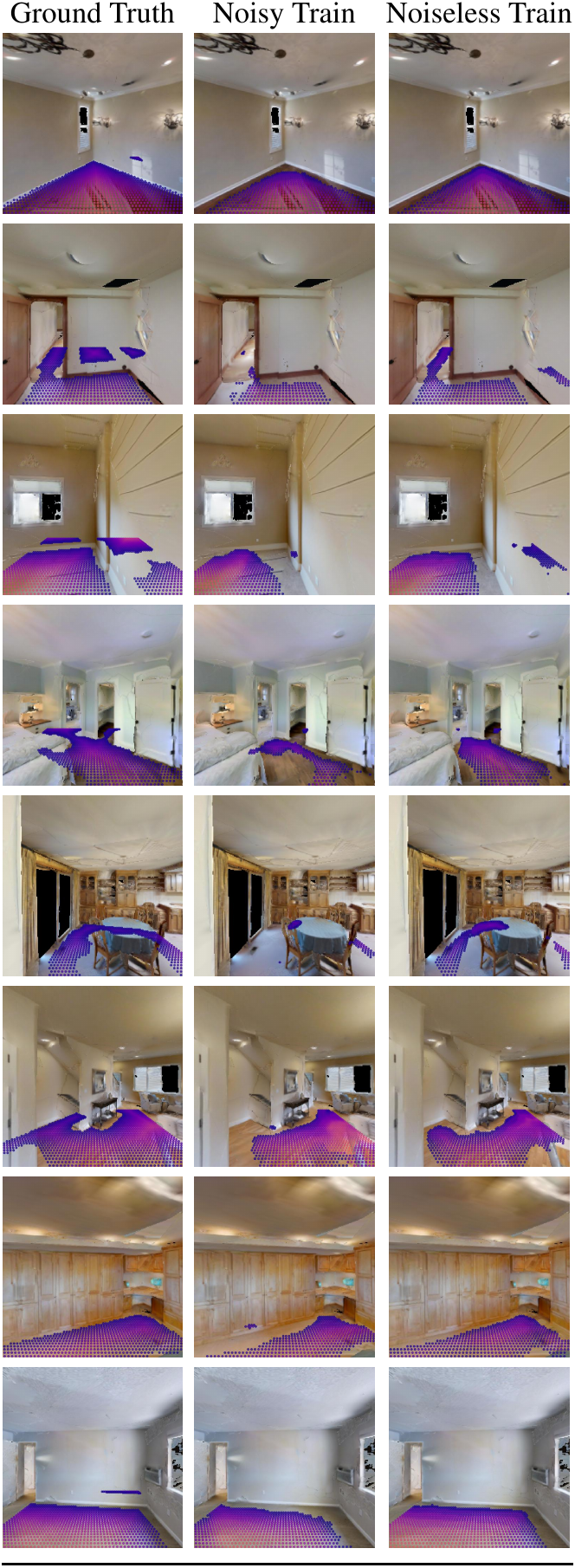}
    \end{minipage}
    \hspace{1em}
    \begin{minipage}{.3\textwidth}
        \includegraphics[width=\linewidth]{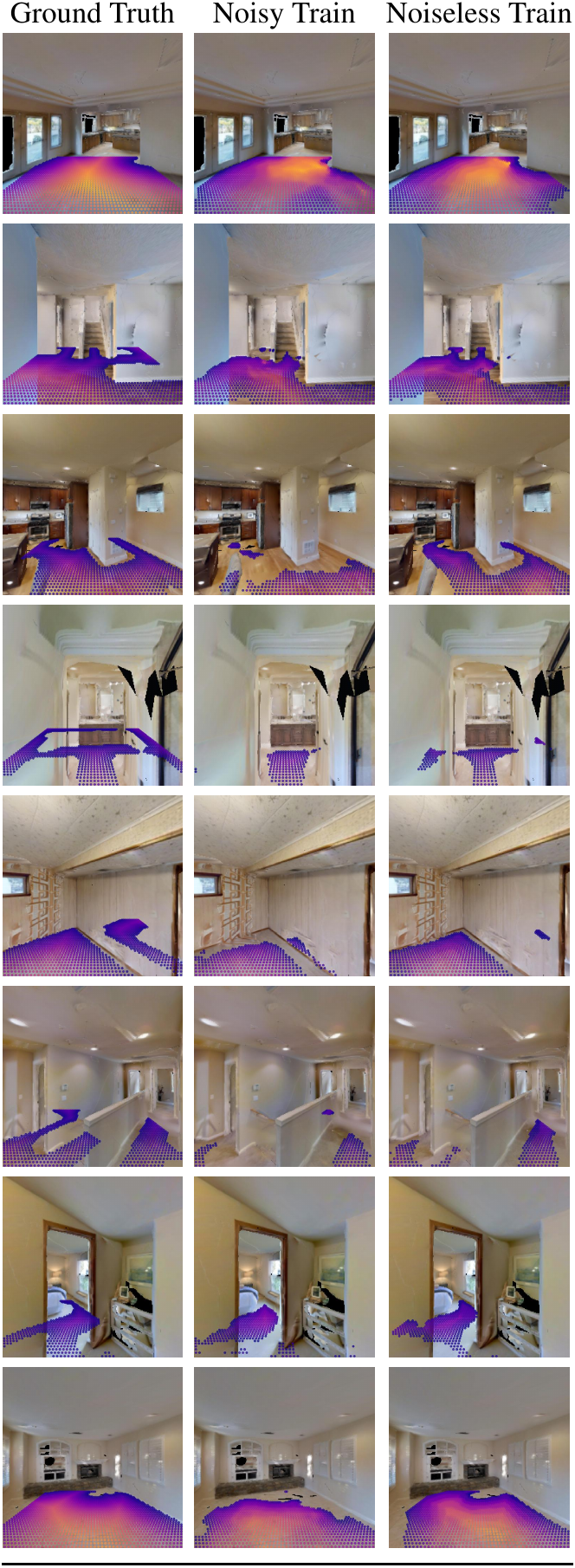}
    \end{minipage}
    \hspace{1em}
    \begin{minipage}{.3\textwidth}
        \includegraphics[width=\linewidth]{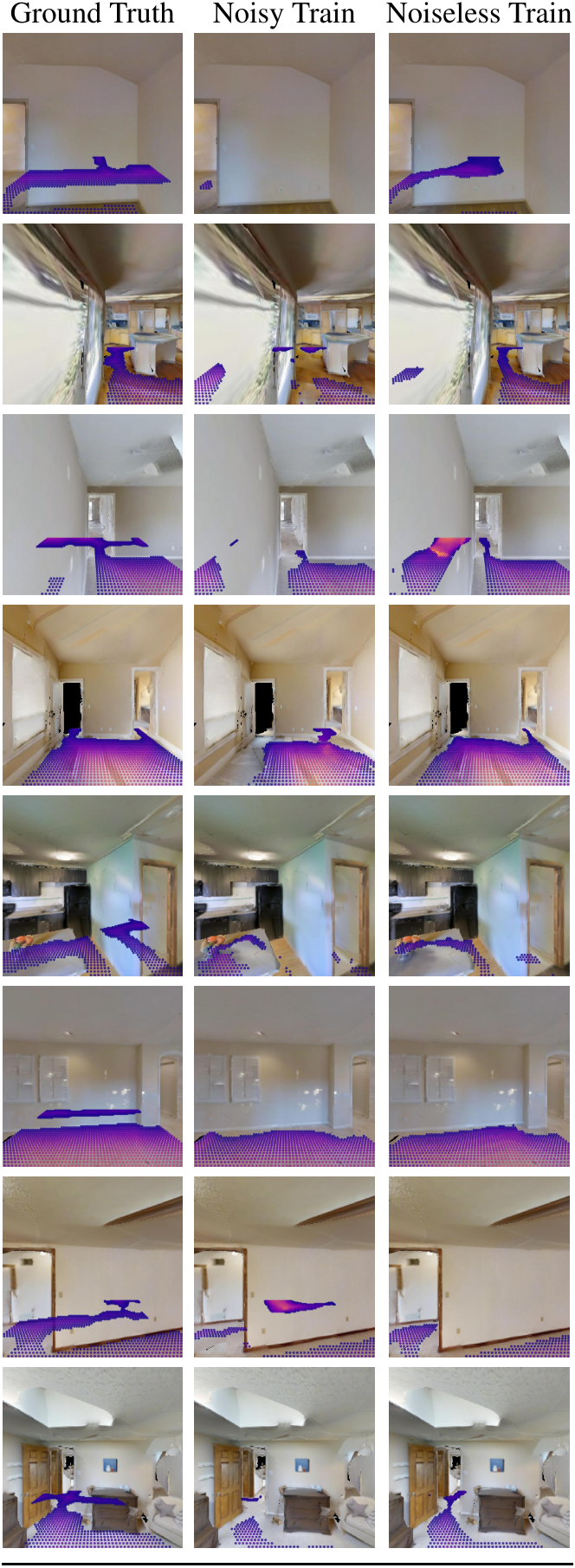}
    \end{minipage}
    \caption{Randomly selected examples of 2D scene distance functions extracted from (left) the simulator navigation-mesh; (middle) a model trained on noisy random walks; or (right) a model trained on noise-free random walks}
    \label{fig:distance_function_supplement}
\end{figure*}

We show extended versions of Figures \ref{fig:timetocollisionalpha} and \label{fig:distance_function} of the main paper in Figures~\ref{fig:angles_supplement} and \ref{fig:distance_function_supplement} respectively. These examples are {\it randomly} selected from the same set of examples used for metric evaluation (examples with at least 10\% of the 4m $\times$ 4m area infront of the agent being freespace).

\clearpage\onecolumn
In ~Fig. \ref{fig:angles_supplement}, we show examples of the conditional probability $P(t | \alpha)$ for grids of scene points and a varying heading angle $\alpha$. This visualizes the underlying representation used to produce distance functions. A distance function can be produced by taking the minimum prediction for each point across all visualized angles. Our method correctly predicts that points for which the heading angle faces a nearby obstacle or surface will have a low distance (shown in purple), whereas points where the heading angle leads to an open space have a higher distance (shown in yellow). In row 2, we see an example of the model's reasoning for various heading angles around a doorway. In the first column, the model recognizes that if the agent is below the doorway in the image it will likely continue into free space, but if the heading angle faces towards the doorway from above it in the image, it is unlikely that the random walk will successfully traverse through, so we see a low prediction. 

\clearpage\onecolumn
In ~Fig. \ref{fig:distance_function_supplement}, we show examples of the final distance function output of a classification model trained with either noisy or noiseless random walks, as compared to the ground truth. As in the main paper, the presence of points is used to indicate whether each method predicted the region to be freespace, so the results support that our model is generally accurate in predicting the visible regions of the scene that are traversable. The model produces lower values near obstacles, and high values in the middle of open spaces.

\clearpage\onecolumn

\begin{figure*}
\centering\includegraphics[scale=0.7]{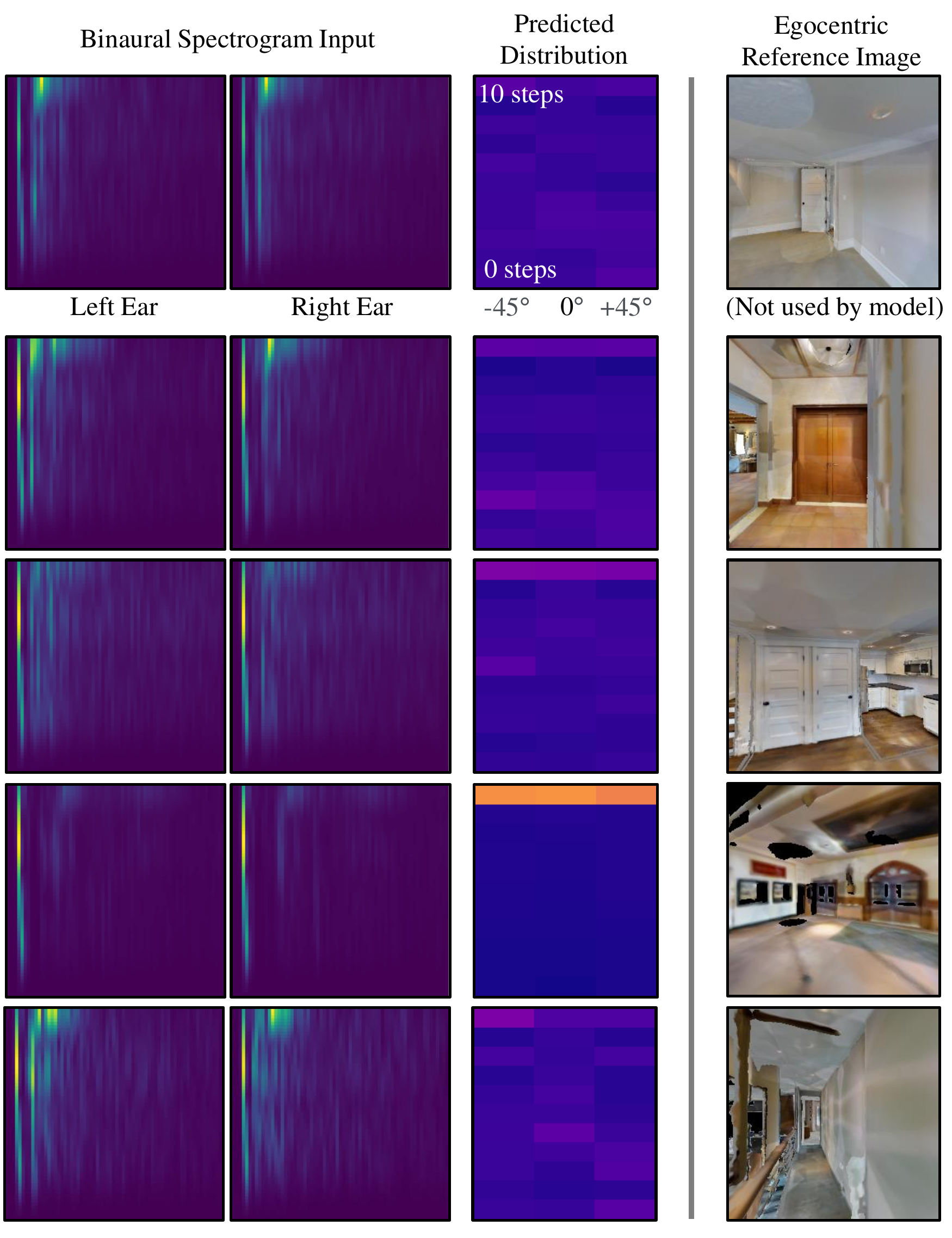} 
\caption{Selected examples of echolocation-based egocentric prediction distributions. Each row shows the model input (left), predicted $P(t|\alpha)$ distribution (middle) and a reference image (right) which was not provided as input to the model.}
\label{fig:echolocation_supplement}
\end{figure*} 

In Fig. \ref{fig:echolocation_supplement} we show examples of per-timestep egocentric predictions made by our egocentric sound-based model from Section \ref{sec:exp_egocentric}. The two spectrogram images shown for each timestep are stacked to create a two channel image input to the model. As with the other egocentric images, the prediction takes the form of a distribution of $P(t | \alpha)$, with angles represented by the turn angle of the action next taken next by the agent.

\end{document}